\documentclass{article}
\usepackage{ijcai23}

\usepackage{times}
\usepackage{soul}
\usepackage{url}
\usepackage[hidelinks]{hyperref}
\usepackage[utf8]{inputenc}
\usepackage[small]{caption}
\usepackage{graphicx}
\usepackage{amsmath}
\usepackage{amsthm}
\usepackage{booktabs}
\usepackage{algorithm}
\usepackage{algorithmic}
\usepackage[switch]{lineno}
\DeclareMathOperator*{\argmax}{arg\,max}

\usepackage{amsfonts}
\usepackage{amsthm}
\usepackage{booktabs}
\usepackage{algorithm}
\usepackage{algorithmic}
\usepackage{tikz}
\usepackage{forest}
\usepackage{bbm}
\usepackage{amsfonts}
\usepackage{booktabs}
\usepackage{algorithm}
\usepackage{algorithmic}
\usepackage{multirow}
\usepackage{subfig}
\newtheorem{definition}{Definition}
\usepackage{amsmath}

\usepackage{makecell}
\usepackage{comment}
\usetikzlibrary{trees,positioning,shapes,shadows,arrows}

\newcommand{\zongxiong}[1]{\textcolor{orange}{{\it [Zongxiong says: #1]}}}

\title{A Comprehensive Study on Dataset Distillation: Performance, Privacy, Robustness and Fairness}

\author{
    Author Name
    \affiliations
    Affiliation
    \emails
    email@example.com
}

\iftrue
\author{
${\textbf{Zongxiong Chen}^*}^2$\and
${\textbf{Jiahui Geng}^*}^1$\and
${\textbf{Derui Zhu}^3}$
${\textbf{Herbert Woisetschläger}}^3$\and 
${\textbf{Qing Li}^1}$\and
${\textbf{Sonja Schimmler}}^2$\and 
${\textbf{Ruben Mayer}}^{3}$ \and
${\textbf{Chunming Rong}}^{1}$ 
\affiliations
$^1$University of Stavanger \\
$^2$ Fraunhofer FOKUS\\
$^3$ Technische Universität München\\
\emails
\{jiahui.geng, chunming.rong, qing.li\}@uis.no, \\
\{derui.zhu, herbert.woisetschlaeger, ruben.mayer\}@tum.de,\\
\{zongxiong.chen, sonja.schimmler\}@fokus.fraunhofer.de
}
\fi

\begin{document}

\maketitle
\def\thefootnote{\textbf{*}}\footnotetext{Equal contributions}

\begin{abstract}
    
    The aim of dataset distillation is to encode the rich features of an original dataset into a tiny dataset. 
    It is a promising approach to accelerate neural network training and related studies. Different approaches have been proposed to improve the informativeness and generalization performance of distilled images. However, no work has comprehensively analyzed this technique from a security perspective and there is a lack of systematic understanding of potential risks. In this work, we conduct extensive experiments to evaluate current state-of-the-art dataset distillation methods. We successfully use membership inference attacks to show that privacy risks still remain. Our work also demonstrates that dataset distillation can cause varying degrees of impact on model robustness and amplify model unfairness across classes when making predictions. This work offers a large-scale benchmarking framework for dataset distillation evaluation.

\end{abstract}
\section{Introduction}
\label{sec:intro}
A recent study~\cite{isenko2022my} has shown that the storage consumption required for popular training sets grows exponentially over time. It is becoming increasingly difficult to train neural networks well on vastly large-scale datasets. Local training is constraint by memory limitations, while distributed training suffers from latency across clients due to network I/O and bandwidth problems. 
Besides, depending on the complexity of the task and model architectures, training deep neural networks requires iterations over the entire training set, up to several hundred or even thousands of times. 
Dataset distillation~\cite{wang2018dataset} may be the holy grail to solve this dilemma. It describes a set of methods aiming to significantly reduce the amount of data while maintaining the same model accuracy level as training on an entire dataset. In contrast to coreset selection, which requires heuristic algorithms and may not be suitable for high-dimensional datasets~\cite{bachem2017practical}, dataset distillation learns to condense large datasets into smaller synthetic datasets rather than using sampling strategies.

There have been many efforts continue to advance dataset distillation, both theoretically and practically~\cite{zhao2021datasetdc,Zhao2021DatasetCW,cazenavette2022dataset,kim2022dataset}. However, most of these works focus on the model's performance and generalization capability across architectures. We state that the security of dataset distillation techniques has been overlooked in the research. Security is essentially critical in the ML application and deployment. Different attacks have been proposed to compromise the integrity of machine learning models. Privacy attacks are trying to infer task-irrelevant private information and restore the training samples from the model's output or parameters~\cite{shokri2017membership,geng2021towards}. There are also attacks on the robustness by adding imperceptible noise to the samples so that the model will be fooled and give false predictions~\cite{xu2020adversarial}. Our work seeks to bridge the gap between dataset distillation and security analysis studies.

In this work, we propose a benchmark to evaluate the security of dataset distillation. Our work focuses on answering the following questions: (1). Is the use of a synthetic dataset instead of the real dataset sufficient to protect data privacy?
(2). What is the impact on the robustness when models are trained on a distilled dataset using fewer training samples with visual noise on each image?
(3). Given that safety-critical applications cannot just focus on the average performance across all classes, is dataset distillation fair for each class in classification tasks?

Based on proposed research questions, we conducted a large-scale analysis of state-of-the-art distillation methods. We use four representative distillation techniques: Differentiable Siamese Augmentation (DSA), Distribution Matching (DM), Training Trajectory Matching (MTT), and Information-Intensive Dataset Condensation (IDC) to synthesize datasets and train neural networks of different architectures on these to obtain the target models. Then we designed a lot of experiments to evaluate the impact of data distillation on the privacy, fairness and robustness of the model. We identify the key factors affecting these metrics by conducting extensive comparative experiments. The experimental results reveal several insightful findings on the dataset distillation, including:
\begin{itemize}
    \item Dataset distillation does amplify the unfairness of the model's predictions between different classes, which increases with the distillation rate.
    \item Dataset distillation does not have natural privacy-preserving capabilities: the success of membership inference attacks will be affected by the distillation rate, initialization, and the number of classes.
    \item The robustness of the model is affected to varying degrees, but the distillation rate plays a minor role here.
\end{itemize}
Here the distillation rate represents the ratio of the numbers of images per class after distillation to before distillation. This is the first work to systematically evaluate the dataset distillation techniques and their risks in terms of security.


\subsection{Outline}
Our paper is organized as follows. We first introduce the background that are related to our work in Section~\ref{sec:background}. Then, we describe the experimental design in detail, including the experimental data, the methods being evaluated, and the evaluation pipeline in Section~\ref{sec:evaluation}. Following that, we present experimental results and key insights 
In Section~\ref{sec:related}, we put our work into context with related works, and conclude with future work in Section~\ref{sec:conclusion}. 


\begin{figure*}[h]
\centering
\subfloat[DM]{\label{fig:dm} \includegraphics[width=0.24\textwidth]{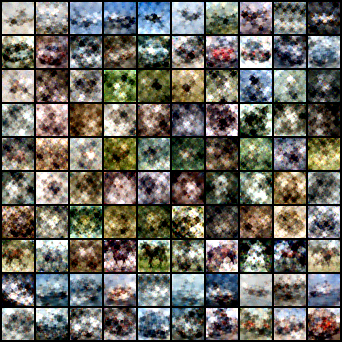}}%
\hfill
\subfloat[DSA]{\label{fig:dsa} \includegraphics[width=0.24\textwidth]{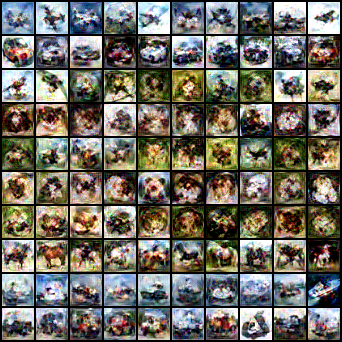}}%
\hfill
\subfloat[MTT]{\label{fig:mtt} \includegraphics[width=0.24\textwidth]{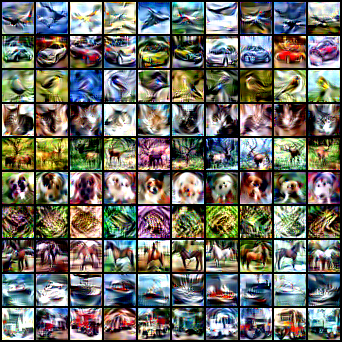}}%
\hfill
\subfloat[IDC]{\label{fig:idc} \includegraphics[width=0.24\textwidth]{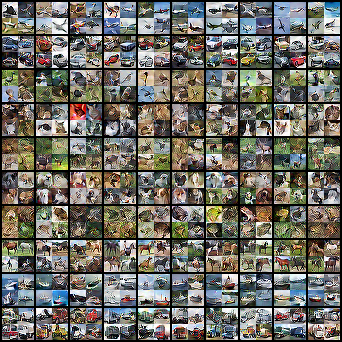}}%
\caption{Distilled 10 images per class from CIFAR10 via DSA, DM, MTT, and IDC. }
\label{fig:synthetic_dataset}
\end{figure*}
\section{Background}
\label{sec:background}

\subsection{Dataset Distillation Basics}
Dataset distillation aims to distill the training set into a much smaller synthetic dataset so that the model trained on the distilled dataset performs as closely as possible to the model trained on the whole dataset. 
Considering the widely used image classification task where the training dataset consists of images and labels $\mathcal{T} = {({x}_i, y_i)}_{i=1}^{\lvert \mathcal{T}\rvert}$. The distilled synthetic dataset what we want to learn is  $\mathcal{S} = {({x}_i, y_i)}_{i=1}^{\lvert \mathcal{S}\rvert}$. Let $\phi_{\theta^{\mathcal{T}}}$ and $\phi_{\theta^{\mathcal{S}}}$ denote the models with parameters $\theta^{\mathcal{T}}$ and $\theta^{\mathcal{S}}$ that trained on $\mathcal{T}$ and $\mathcal{S}$ respectively. 
Therefore, an ideal dataset distillation should satisfy:
\begin{equation}
    \mathbb{E}_{x\sim P_{\mathcal{D}}}[\ell (\phi_{\theta^{\mathcal{T}}}({x}), y)] \simeq \mathbb{E}_{x\sim P_{\mathcal{D}}}[\ell (\phi_{\theta^{\mathcal{S}}}({x}), y)],
\end{equation}
where $\ell$ is the loss function and $P_{\mathcal{D}}$ is the real data distribution.


\subsection{Security of ML Models}

\paragraph{Privacy Perspective}

When the deep model learns high-level and abstract features, it inevitably remembers too many details from the training samples, including a lot of task-independent private information. Many attacks attempt to attack the target model to infer the privacy. Representative attacks include membership inference attacks (MIAs)~\cite{shokri2017membership}, attribute inference attacks~\cite{gong2016you}, model inversion attacks~\cite{fredrikson2015model,zhang2020secret}, gradient inversion attacks~\cite{zhu2019deep}, etc. The goal of MIAs is to discriminate whether a specific data participates in the model training and it can be regarded as a binary classification problem. Model inversion attack and gradient inversion attack attempt to restore training data and labels from the accessed model parameters or gradient parameters. Attribute inference attacks attempt to infer task-independent attributes. 

\paragraph{Robustness Perspective}
 In machine learning, model robustness refers to how well a model can perform on new data compared with training data. Ideally, performance should not deviate significantly.
 However, deep models have recently been shown to be unstable to adversarial perturbations, i.e. even imperceptible perturbations are sufficient to fool the well-trained models and results in incorrect predictions.
\begin{definition}[Adversarial Perturbation] For a given classifier denoted by $f(\cdot; \theta): \mathbb{R}^d \rightarrow \mathbb{R}^K$, with input $x \in \mathbb{R}^d$ and $K$ is the number of classes in the task. Let $f(x; \theta)$ represent the model softmax output and $\hat{f}(x; \theta)$ represent the index of the highest probability. Then, the adversarial perturbation is defined as the minimum perturbation $r$ that is sufficient to change the model prediction:
\begin{equation}
    \Delta(x; f(\cdot; \theta)):= \min_{r} \lVert r \rVert_2 \quad s.t. \quad \hat{f}(x+r) \neq \hat{f}(x)
\end{equation}
\end{definition}
Adversarial accuracy is a commonly used method to measure the robustness of a classifier. Usually, we use different adversarial sample generation algorithms to add a perturbation of magnitude $\epsilon$ to the sample and test the accuracy of the model.
\begin{definition}[Adversarial Accuracy]
Let $\mathcal{L}(f(x; \theta), y)$ be the classification error given input $x$, classifier $f(\cdot;\theta)$ and true label $y$. Adversarial accuracy an is defined as follows for an adversary with an adversary region $R(x)$. Commonly, we consider $\mathbb{B}(x, \epsilon)$, a $\epsilon-$ball around $x$ based on the distance metric $d$ as the adversary region $R(x)$.
\begin{equation}
    acc = \mathbb{E}_{(x, y) \sim \mathcal{D}} [\mathbbm{1}(f(x^*; \theta) = y)]
\end{equation}
where 
\begin{equation}
    x^* = \argmax\limits_{x' \in R(x)} \mathcal{L}(x', y).
\end{equation}
\end{definition}


\paragraph{Fairness Perspective}
Fairness between classes is also an important indicator of safety. Often people only report average performance, i.e., average metrics across classes, which can lead to an overestimation of safety. Fine-grained performance per category is essential for safety-critical applications, such as the 'human' class in autonomous driving. It has been found that many post-processed machine learning models amplify the unfairness between different categories, i.e., the models perform well on some categories while degrading a lot on others. Hooker et al.~\cite{hooker2020characterising} show that model compression can amplify model bias in models, including using model pruning and quantization protocols. Silva et al. ~\cite{silva2021towards} find that distilled models almost always exhibit statistically significant bias compared to the original pre-trained models. ~\cite{ma2022tradeoff,benz2021robustness,xu2021robust} shows that models trained with adversarial examples exhibit the same problem. 
Inspired by \cite{li2021ditto,lin2022personalized}, we give the definition of fairness as follows, 

\begin{definition}[Performance Fairness]\label{def:performance-fairness}
A model $w_1$ is more fair than $w_2$ if the test performance distribution of $w_1$ is more uniform than that of $w_2$, e.g. $\text{var}({F(w_1)}) < \text{var}({F(w_2)})$ , where $F(*)$ denotes evaluation metrics such as test accuracy or test loss of model among all classes and $\text{var}$ denotes the variance. 
\end{definition}
In contrast to the definition in \cite{li2021ditto} which focuses on comparison in federated learning settings, our definition can be applied to evaluate the fairness about different models directly. 

\section{Evaluation Benchmark}
\label{sec:evaluation}
Our work unites data distillation and privacy, robustness and fairness in ML research.  For that, we introduce evaluation pipelines along with four state-of-the-art dataset distillation methods.

\subsection{Dataset}
We implement our evaluation pipeline on CIFAR10 and CIFAR100, which are widely used for image classification tasks. CIFAR-10 contains 10 different categories, each category has 60K (50K training images and 10K test images) images of size $32\times 32$. CIFAR-100 contains 100 different categories, each with 500 training images and 100 test images of the same size. Using larger datasets is a trend of dataset distillation, such as TinyImageNet~\cite{Le2015TinyIV}, ImageNet~\cite{ILSVRC15} and CelebA~\cite{liu2015faceattributes}, which is also our future work.

 \subsection{Dataset Distillation Methods}
We choose DSA, DM, MTT, IDC as representative methods to be evaluated in our benchmark. They are all accepted by top conferences with code public released. They have been widely compared in different works.
 \begin{itemize}
     \item \textbf{DSA} \\
     DSA is essentially based on the gradient matching mechanism, which infers the synthetic dataset by matching the model optimization trajectory trained on the synthetic dataset with the trajectory trained on the original dataset. DSA proposed to apply Differentiable Siamese Augmentation during the optimization. To make it applicable to any initial state, the optimization objective of the model can be expressed as:

\begin{equation}
    \tiny
    \min_{\mathcal{S}} \mathbb{E}_{\theta_0 \sim P_{\theta_0}} [\sum_{t=1}^{T-1} D(\nabla_{\theta} \mathcal{L}(\mathcal{A}(\mathcal{S}, \omega^{\mathcal{S}}), \theta_t), \nabla_{\theta} \mathcal{L}(\mathcal{A}(\mathcal{T}, \omega^{\mathcal{T}}), \theta_t))]
\end{equation}
where $T$ is the number of iterations, $\theta_0$ is the random initialized parameters, and  $\mathcal{A}$ is family of image transformations that parameterized with $\omega^{\mathcal{S}}$ and $\omega^{\mathcal{T}}$ for the synthetic and real training samples respectively.
     
     
     \item \textbf{DM} \\
     DM proposes to optimize the distilled dataset by reducing the maximum mean discrepancy (MMD) between the synthetic and real data. DM uses neural network with different random initializations instead of sampling parameters from a set of pretrained networks. In addition, DM also uses Differentiable Siamese Augmentation. The objective function can be illustrated as:
\begin{equation}
\label{eq:dm}
\tiny
    \min_{\mathcal{S}} \mathbb{E}_{\theta \sim P_{\theta}\atop \omega \sim \Omega} \lVert \frac{1}{\lvert \mathcal{S} \rvert} \sum_{i=1}^{\lvert \mathcal{S} \rvert} \psi_{\theta}(\mathcal{A}(s_i, \omega)) - \frac{1}{\lvert \mathcal{T} \rvert} \sum_{i=1}^{\lvert \mathcal{T} \rvert} \psi_{\theta}(\mathcal{A}(x_i, \omega))\rVert^2,
\end{equation}
where $\psi_{\theta}$ is a family of deep models  to map the samples into a lower dimensional space, and $\omega \sim \Omega$ is the augmentation parameter. Importantly, Equation~\ref{eq:dm} requires only optimizing the synthetic dataset $\mathcal{S}$, not the model parameters $\theta$.
     \item \textbf{MTT} \\
     MTT regards the sequence of models trained on the real data as the expert trajectory, and the model trained on the synthetic data as the student model. It first samples $\theta_t$ at a random timestep $t$ from the expert trajectory to initialize the student model. Then it performs N Step gradient descent updates on the student model and selects the expert model $\theta_{n+M}$ as the target model. The objective of MTT is to make the student model approach the expert model and use the Euclidean distance of the expert parameters to normalize the matching loss: 
    \begin{equation}
        \mathcal{L} = \frac{\lVert \hat{\theta}_{t+N} - \theta^*_{t+M} \lVert_2^2}{\lVert \hat{\theta}_t^* - \theta^*_{t+M} \lVert_2^2}
    \end{equation}
     \item \textbf{IDC} \\
   Similar to DSA, IDC is also use gradient matching algorithm to obtain optimal distlled dataset. The key difference between DSA and IDC is that IDC introduces a multi-formation function: $f:\mathbb{R}^{n\times m} \rightarrow \mathbb{R}^{n'\times m}$, which can synthesize more informative dataset under the same storage consumption. The multi-formation function $f$ can be either learned or a pretrained model during optimization. 

    
 \end{itemize}

\subsection{Evaluation Pipeline}
Our evaluation pipeline is mainly divided into three stages. The first stage is image distillation, we use the Github codes of DSA\footnote{\url{https://github.com/VICO-UoE/DatasetCondensation}}, DM\footnote{\url{https://github.com/VICO-UoE/DatasetCondensation}}, MTT\footnote{\url{https://github.com/GeorgeCazenavette/mtt-distillation}}, and IDC \footnote{\url{https://github.com/snu-mllab/Efficient-Dataset-Condensation}} official releases to obtain different distilled datasets. The second stage is to use the distilled data set to train a randomly initialized model. The third stage is to design targeted experiments to analyze the target model. We evaluate target models in term of performance, privacy, robustness and fairness. As in prior works, the informativeness of the distilled image and the generalization performance are two fundamental metrics to measure the performance. Image informativeness is usually measured by the accuracy of the neural network trained on it, and a higher accuracy of the model means a better representation of the image. Furthermore, generalization is defined as the model accuracy across different architectures. We chose to use membership inference attacks among the many ways to attack the privacy of machine learning models. Model inversion attacks and gradient inversion attacks are not suitable for evaluating the privacy-preserving capability of dataset distillation due to the irreversible information loss after distillation. We perform adversarial attacks on this well-trained model to analysis the intrinsic robustness issues caused by the distilled dataset. In our evaluation, we consider the degradation of the robust accuracy caused by DeepFoolAttack~\cite{moosavi2016deepfool} overall results across models .As the fairness study, we observe how the model's accuracy and loss vary across different classes, and whether the model has a prominent drop on spesific classes.

\label{sec:fair}

\begin{figure*}[h]
\centering
\subfloat[DSA]{\label{fig:tsne-dsa} \includegraphics[angle=0,trim={2cm 0.2cm 2cm 0.2cm},clip,width=0.23\textwidth]{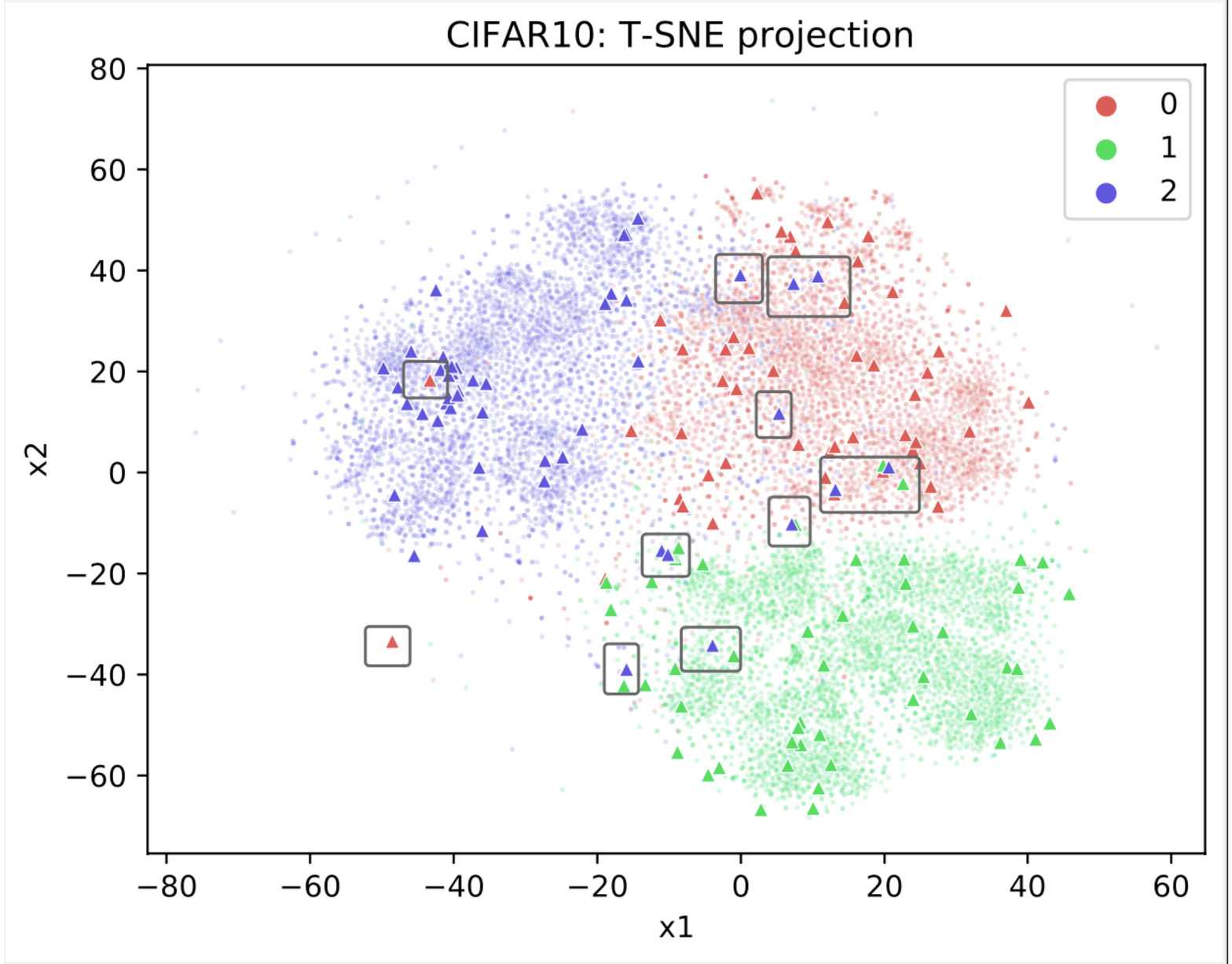}}%
\hfill
\subfloat[DM]{\label{fig:tsne-dm} \includegraphics[angle=0,trim={2cm 0.2cm 1cm 0.4cm},clip,width=0.24\textwidth]{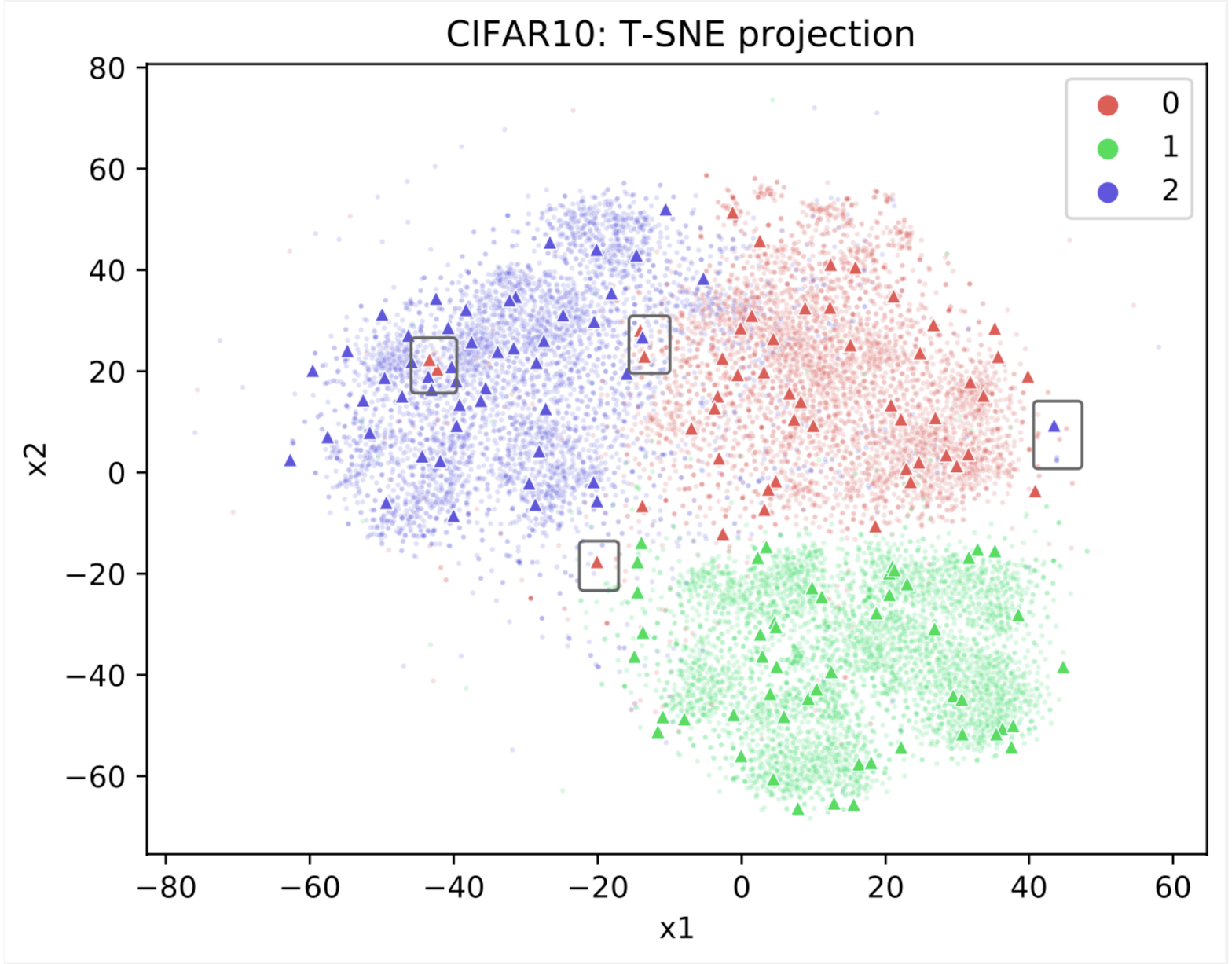}}%
\hfill
\subfloat[MTT]{\label{fig:tsne-mtt} \includegraphics[angle=0,trim={2cm 0.2cm 1cm 0.4cm},clip,width=0.24\textwidth]{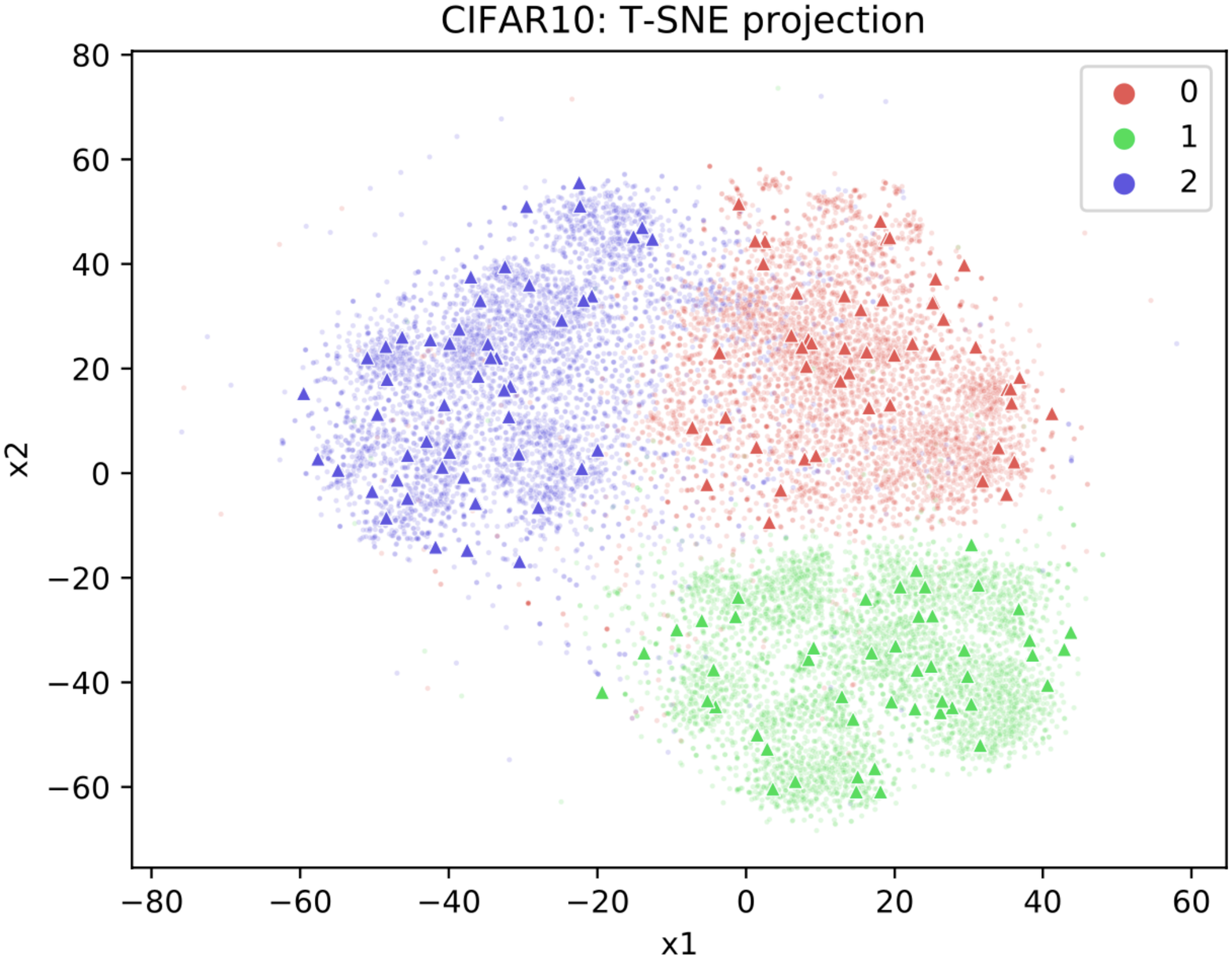}}%
\hfill
\subfloat[IDC]{\label{fig:tsne-idc} \includegraphics[angle=0,trim={2cm 0.7cm 2cm 0.4cm},clip,width=0.24\textwidth]{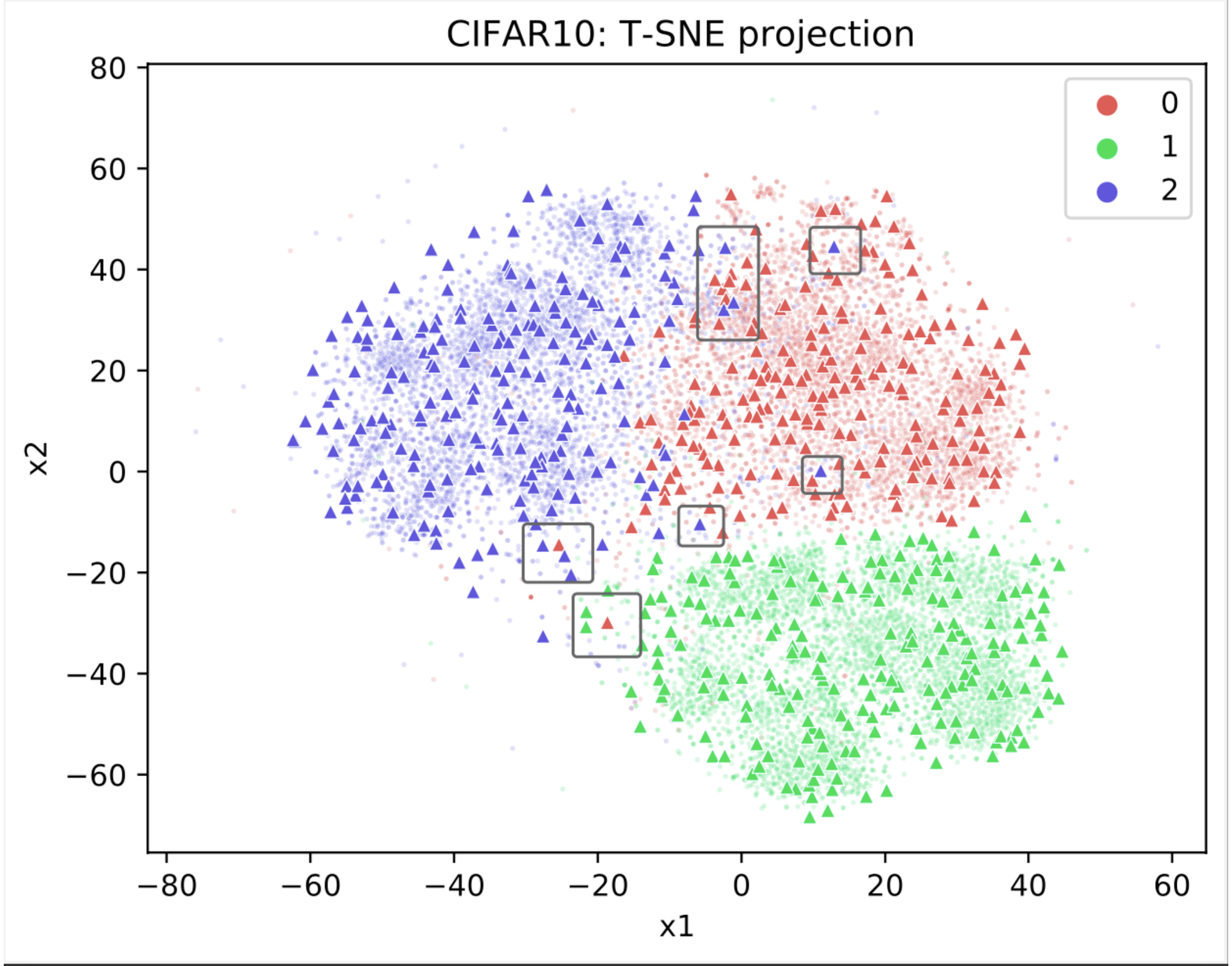}}%
\caption{Data distribution of real images and synthetic images learned in CIFAR10 at IPC-50. We use a pretrained ConvNet3 model to generate an 2048-dimension embedding and project that embedding into 2-dimension plane by t-SNE with hyperparameter \textit{perplexity} 50. The black boxs in presented in the figure indicate the obvious outliers in the projected embeddings.}
\label{fig:tsne}
\end{figure*}

\begin{figure*}[htb]
\centering
\subfloat[IPC=50]{\label{fig:robustall} \includegraphics[width=0.19\textwidth]{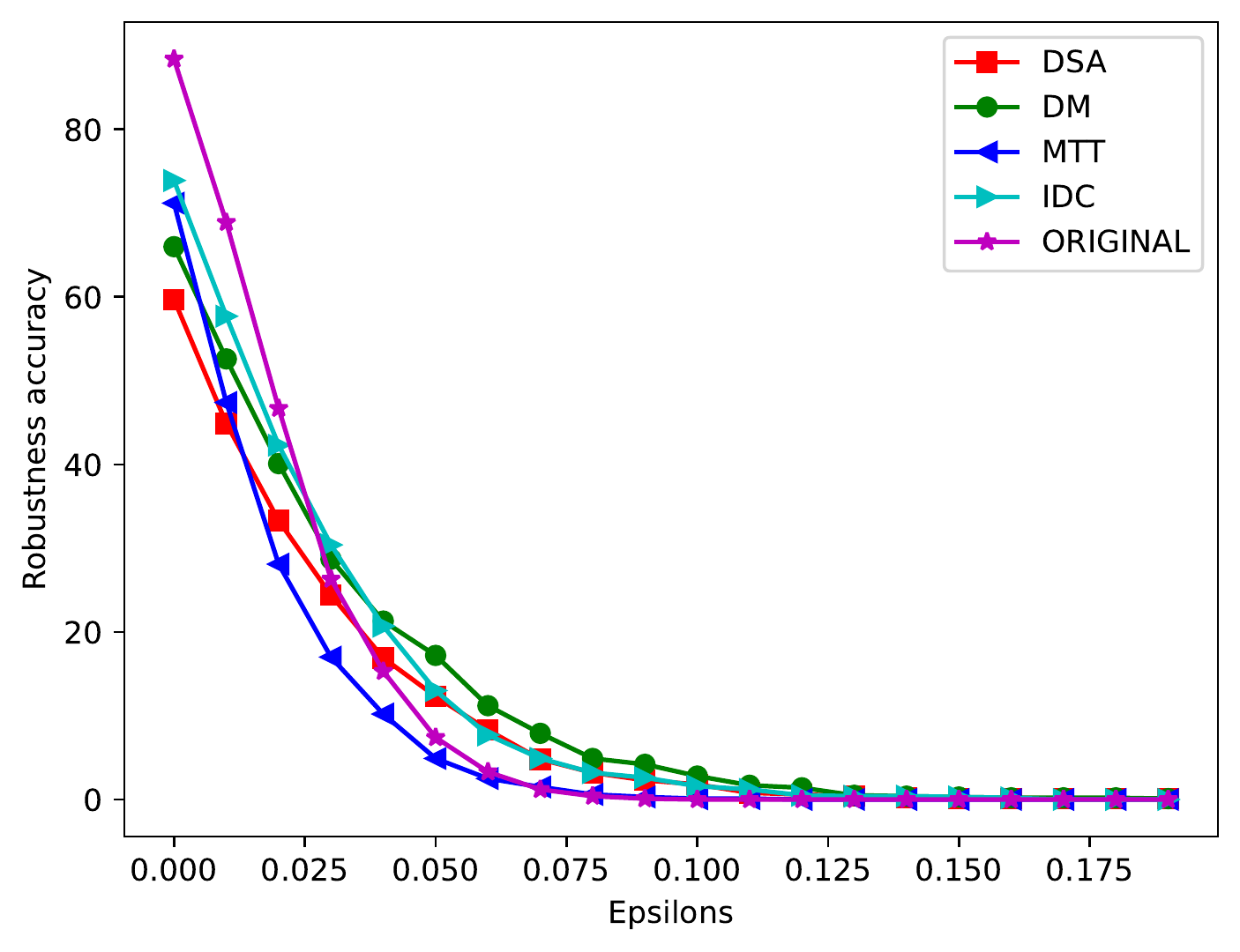}}%
\hfill
\subfloat[DSA]{\label{fig:robustdsa} \includegraphics[width=0.19\textwidth]{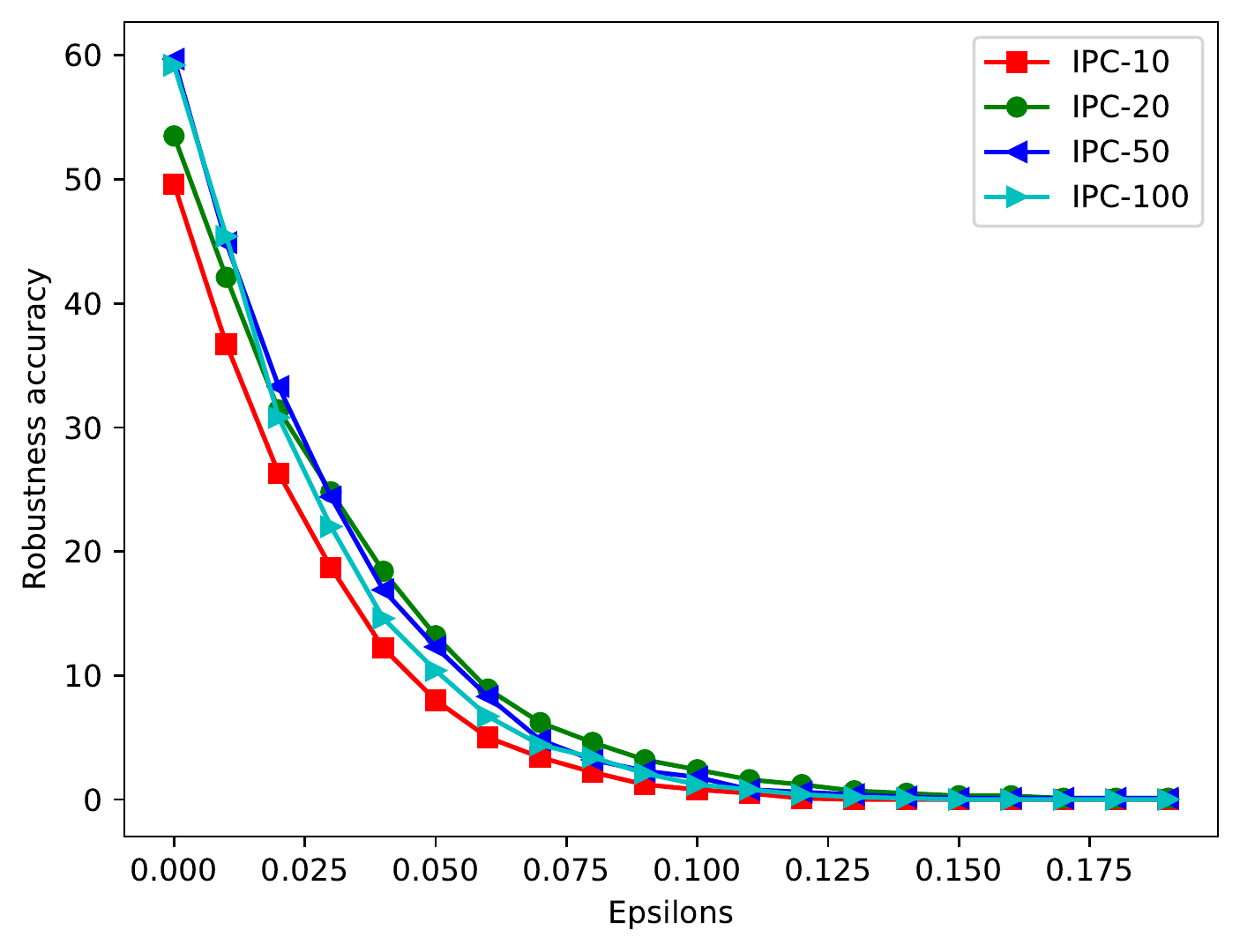}}%
\hfill
\subfloat[DM]{\label{fig:robustdm} \includegraphics[width=0.19\textwidth]{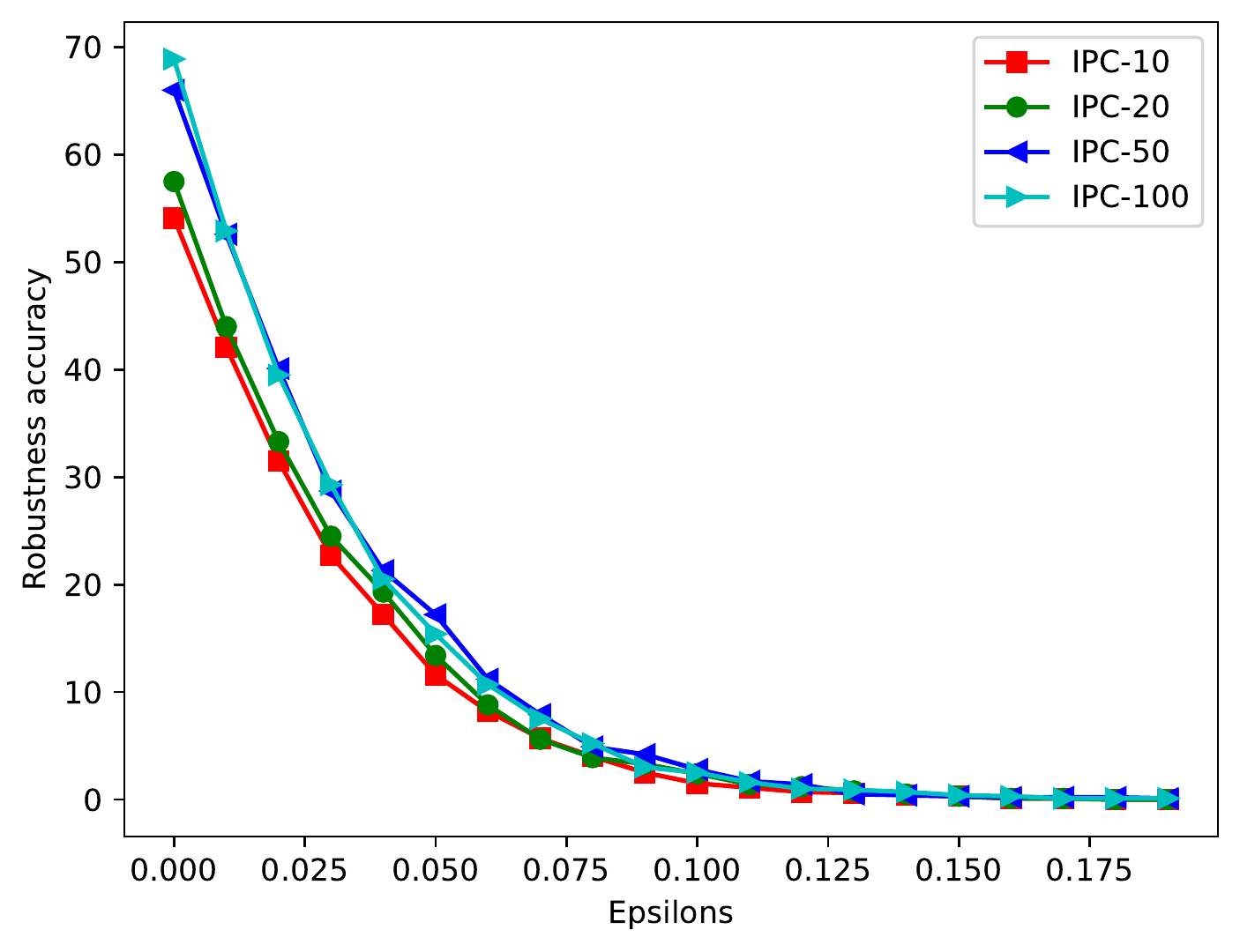}}%
\hfill
\subfloat[MTT]{\label{fig:robustmtt} \includegraphics[width=0.19\textwidth]{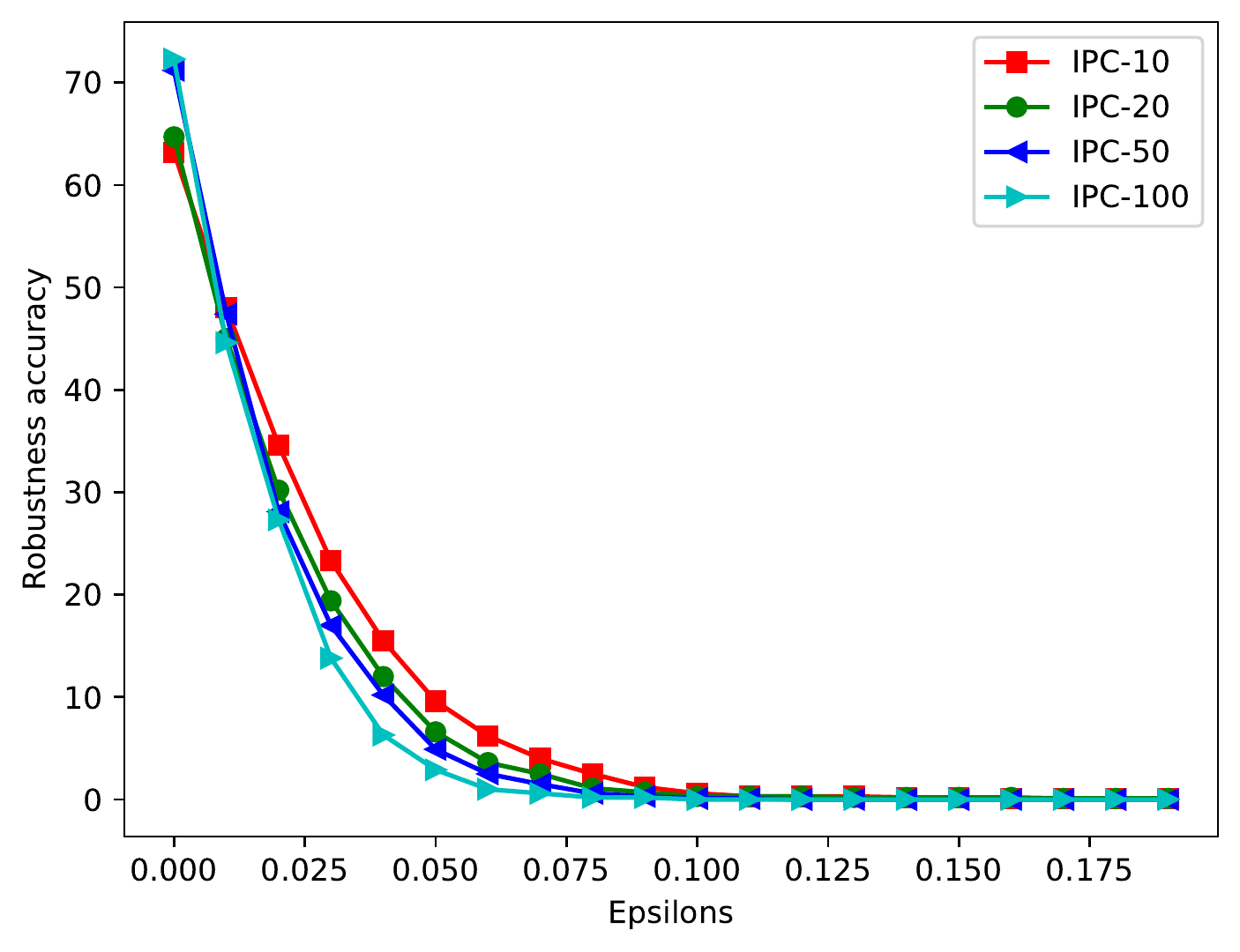}}%
\hfill
\subfloat[IDC]{\label{fig:robustidc} \includegraphics[width=0.19\textwidth]{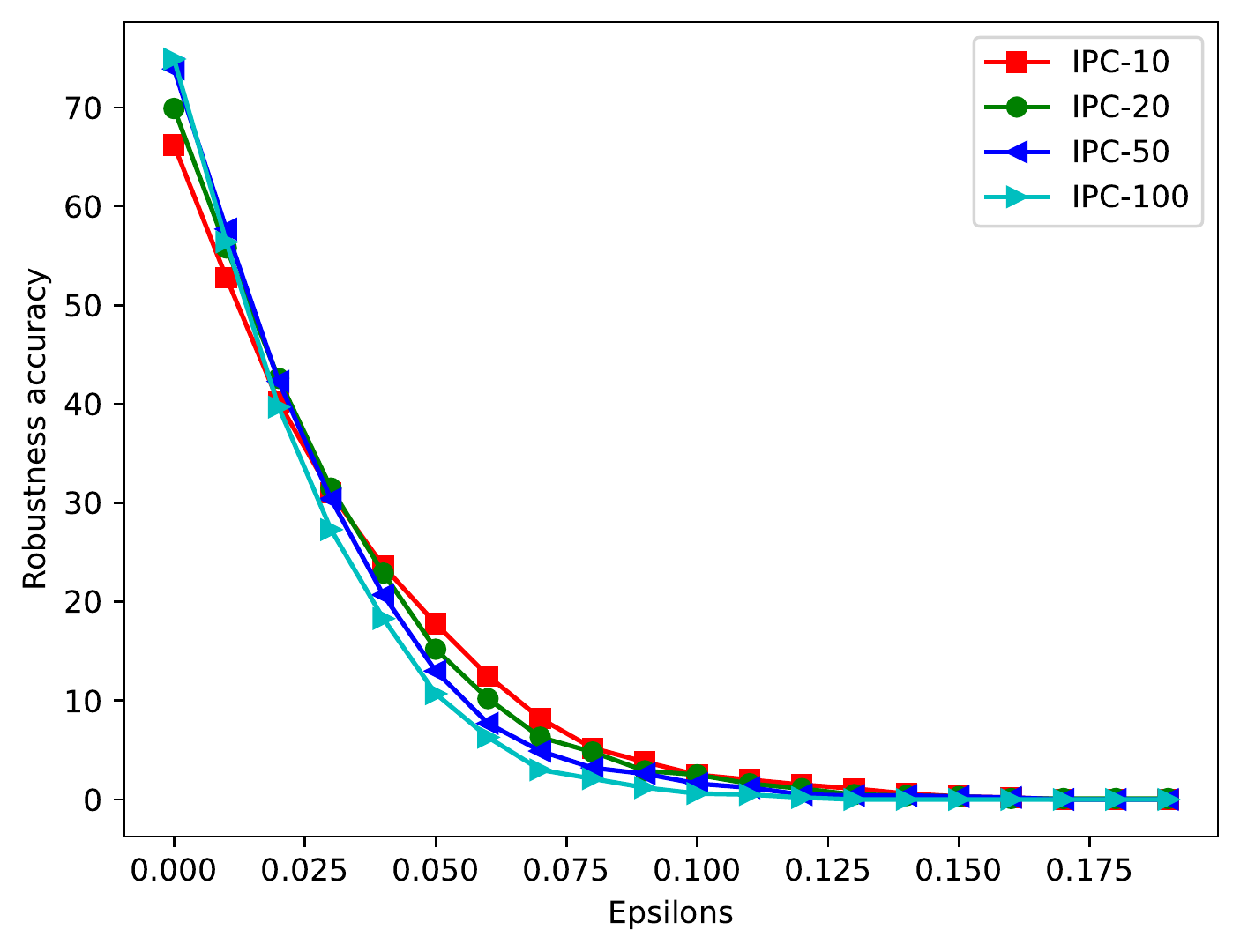}}%
\caption{Robustness Evaluation on CIFAR10}
\label{fig:all-robustallcifar10}
\subfloat[IPC=50]{\label{fig:robustallcifar100} \includegraphics[width=0.19\textwidth]{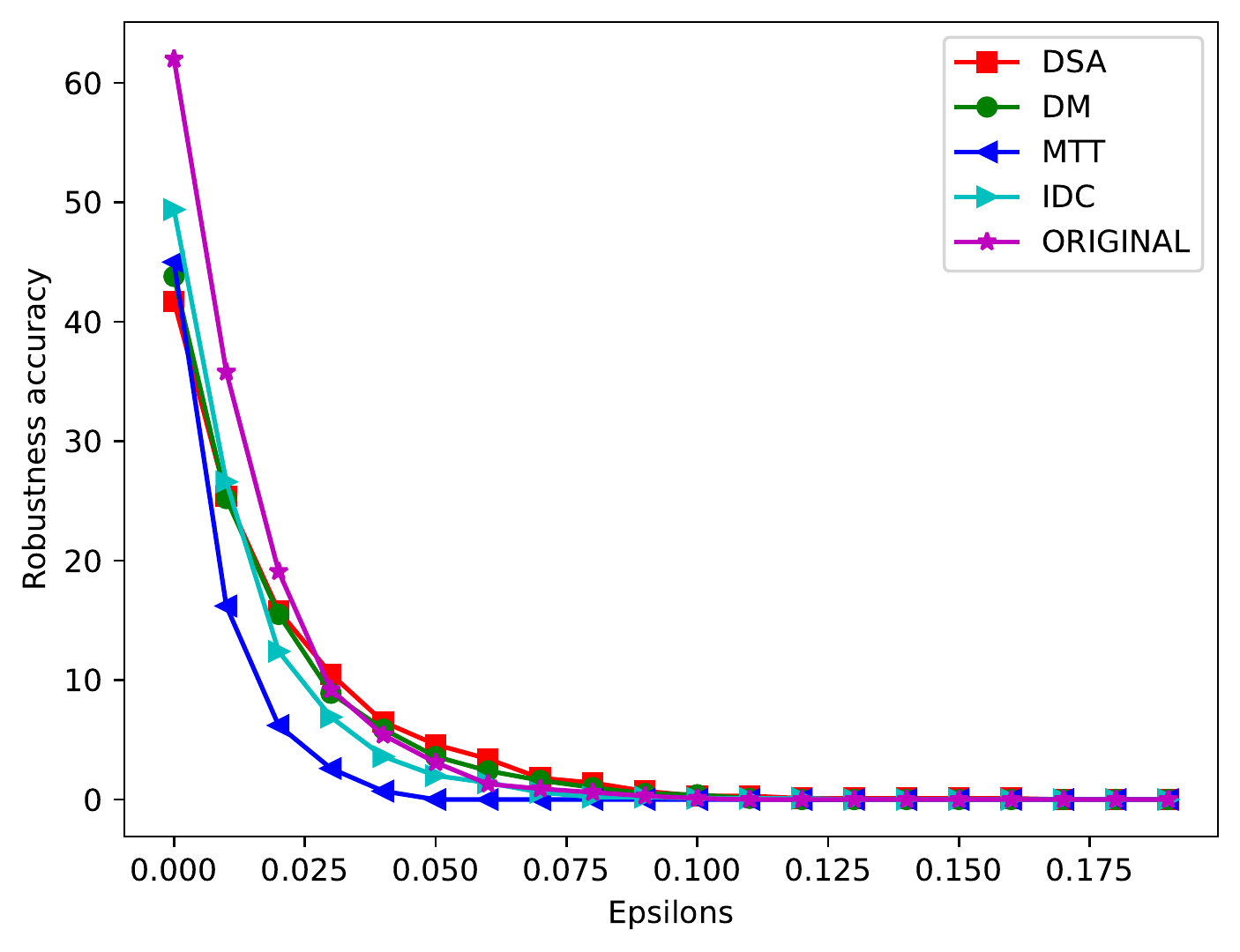}}%
\hfill
\subfloat[DSA]{\label{fig:robustdsa} \includegraphics[width=0.19\textwidth]{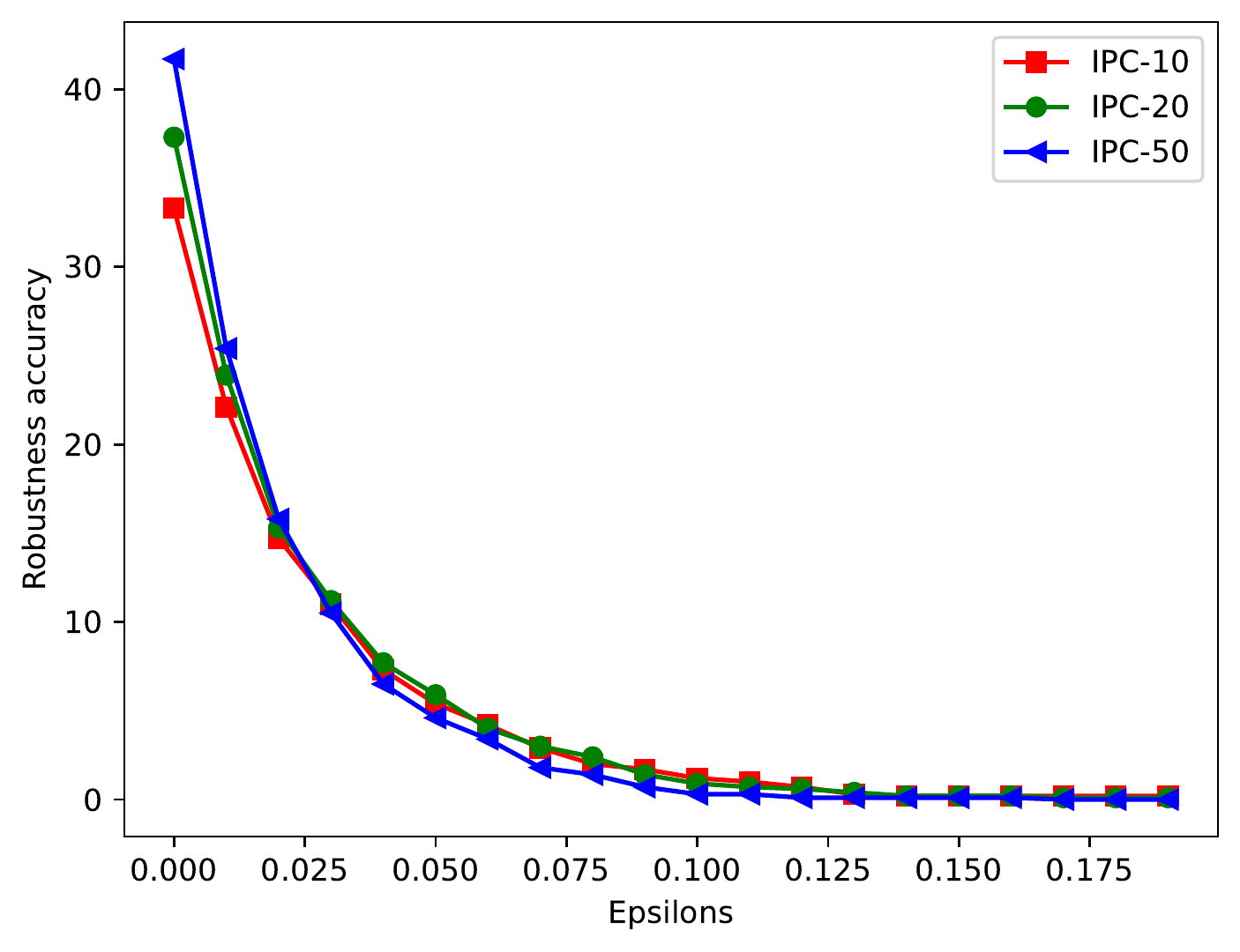}}%
\hfill
\subfloat[DM]{\label{fig:robustdm} \includegraphics[width=0.19\textwidth]{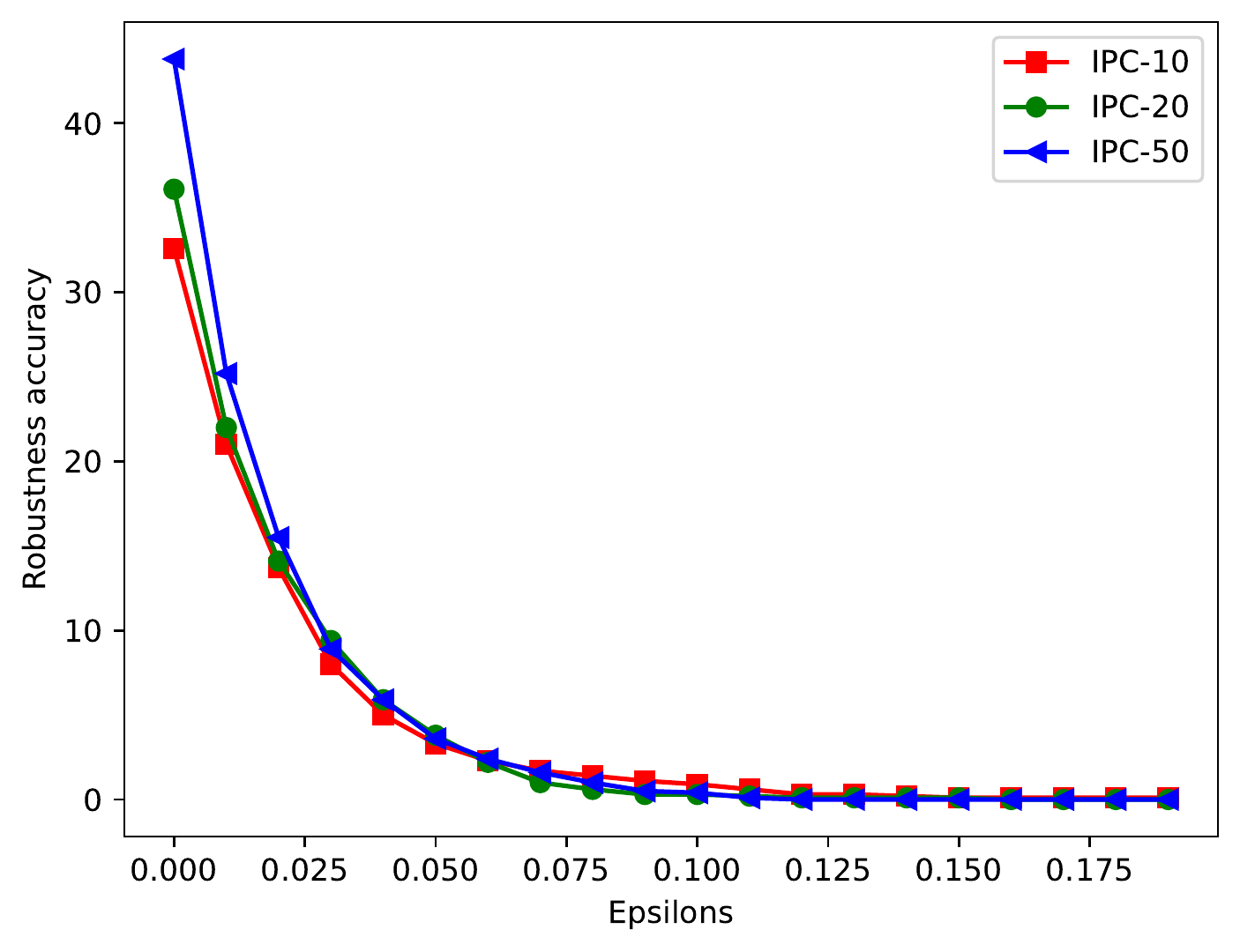}}%
\hfill
\subfloat[MTT]{\label{fig:robustmtt} \includegraphics[width=0.19\textwidth]{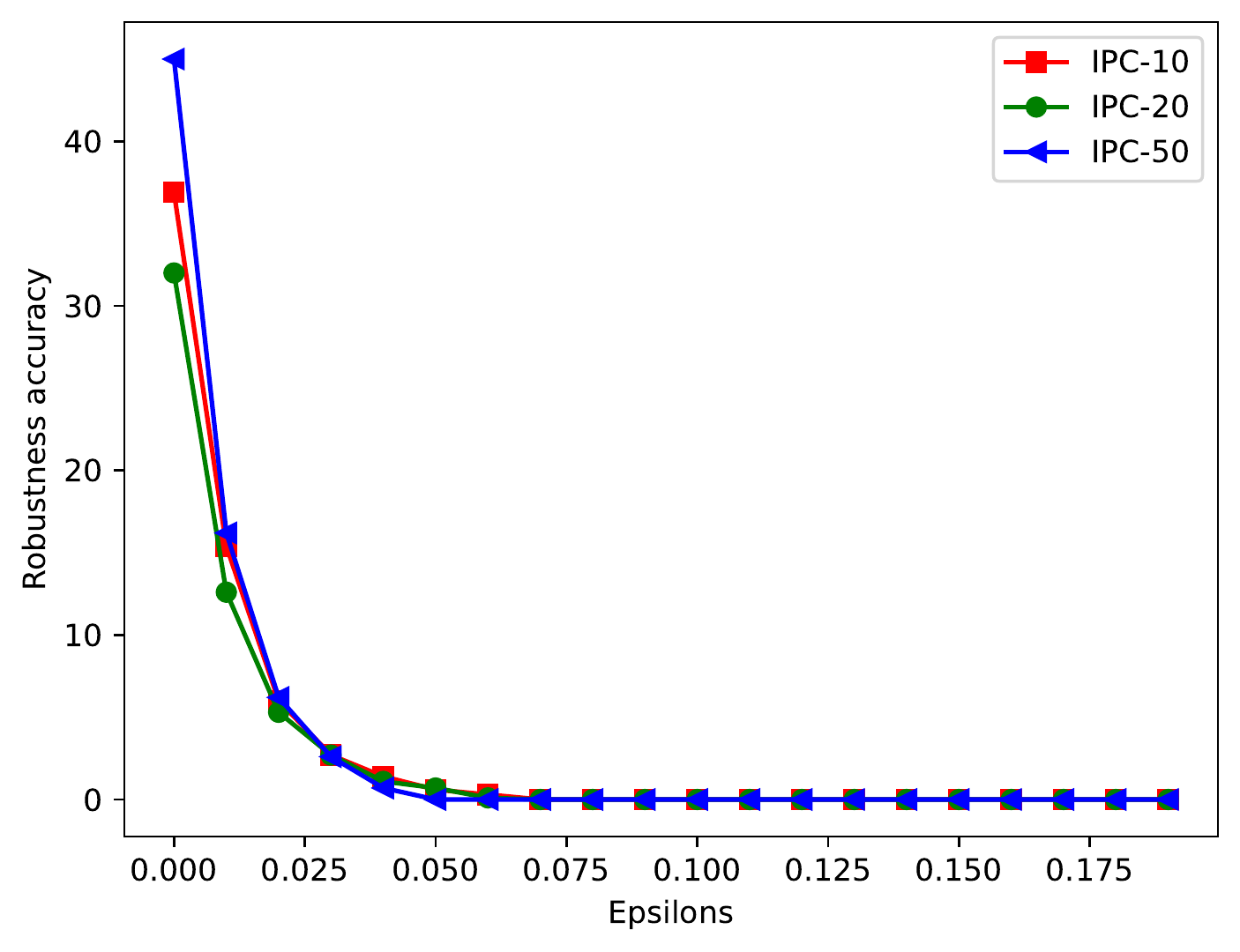}}%
\hfill
\subfloat[IDC]{\label{fig:robustidc} \includegraphics[width=0.19\textwidth]{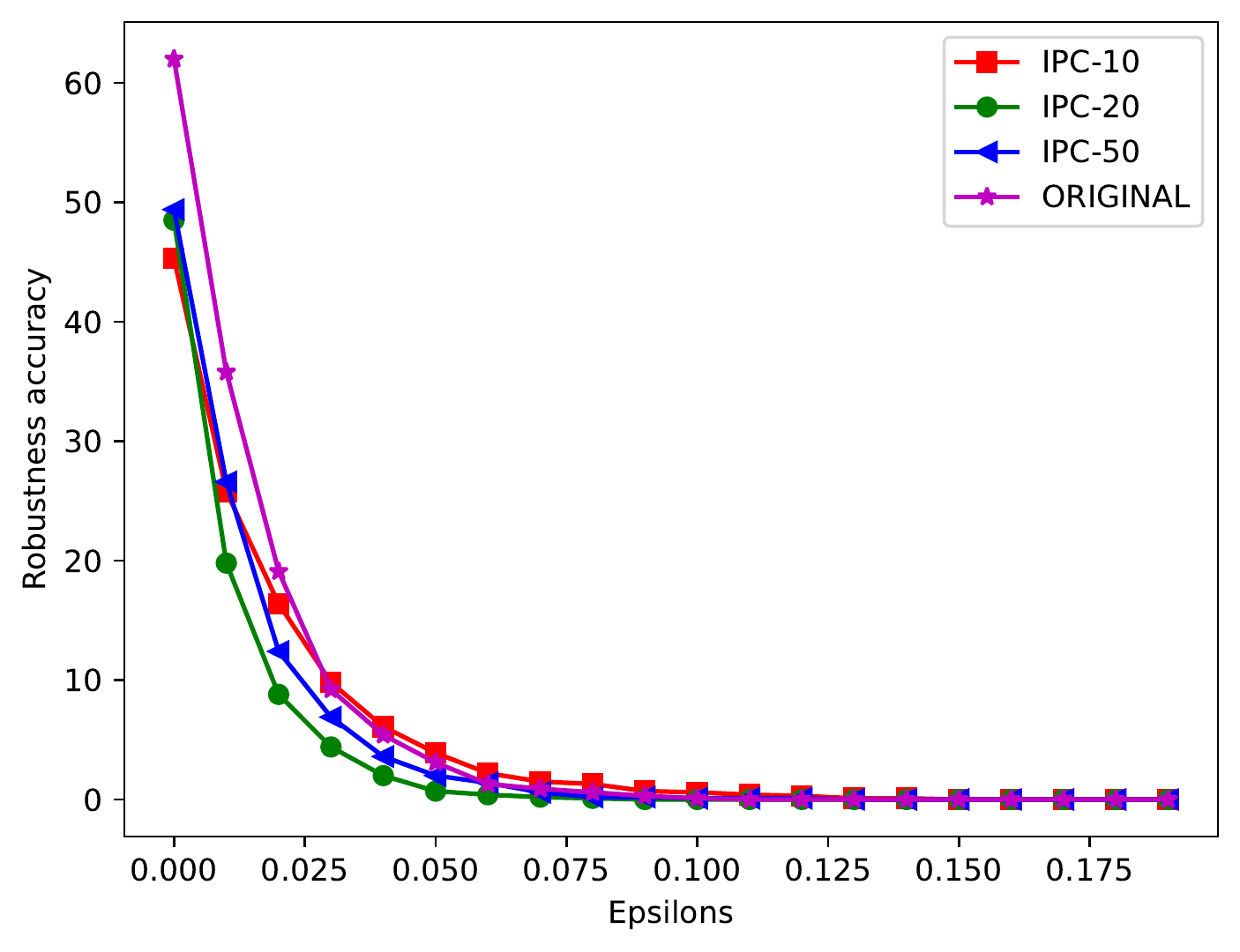}}%
\caption{Robustness Evaluation on CIFAR100}
\label{fig:all-robustallcifar100}
\label{5figs}

\end{figure*}
\section{Experimental Studies}
\label{sec:experiments}

\begin{figure*}[h]
\centering
\subfloat[CIFAR10]{\label{fig:cifar10acc}\includegraphics[width=0.22\textwidth]{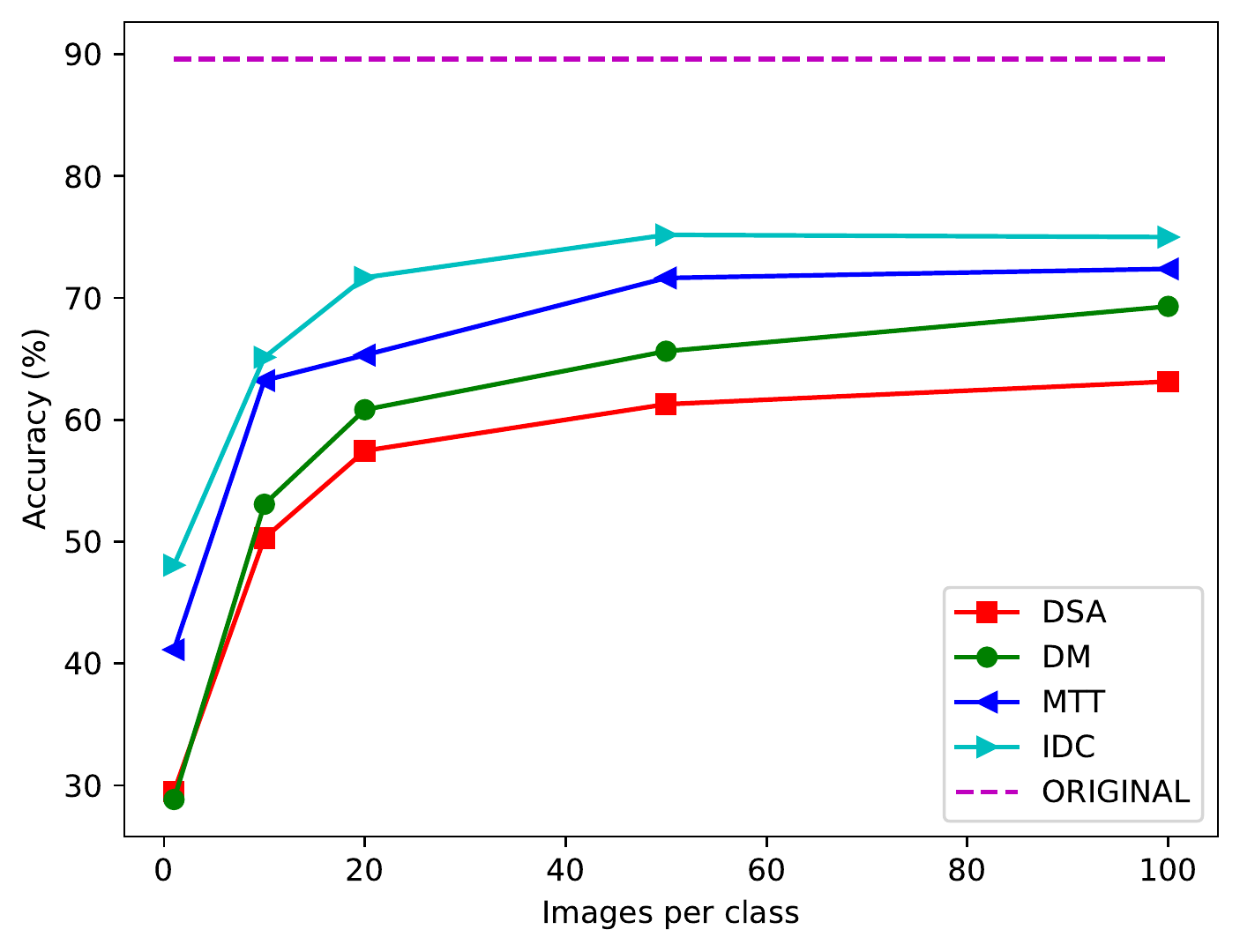}}\hfil
\subfloat[CIFAR100]{\label{fig:cifar100acc}\includegraphics[width=0.22\textwidth]{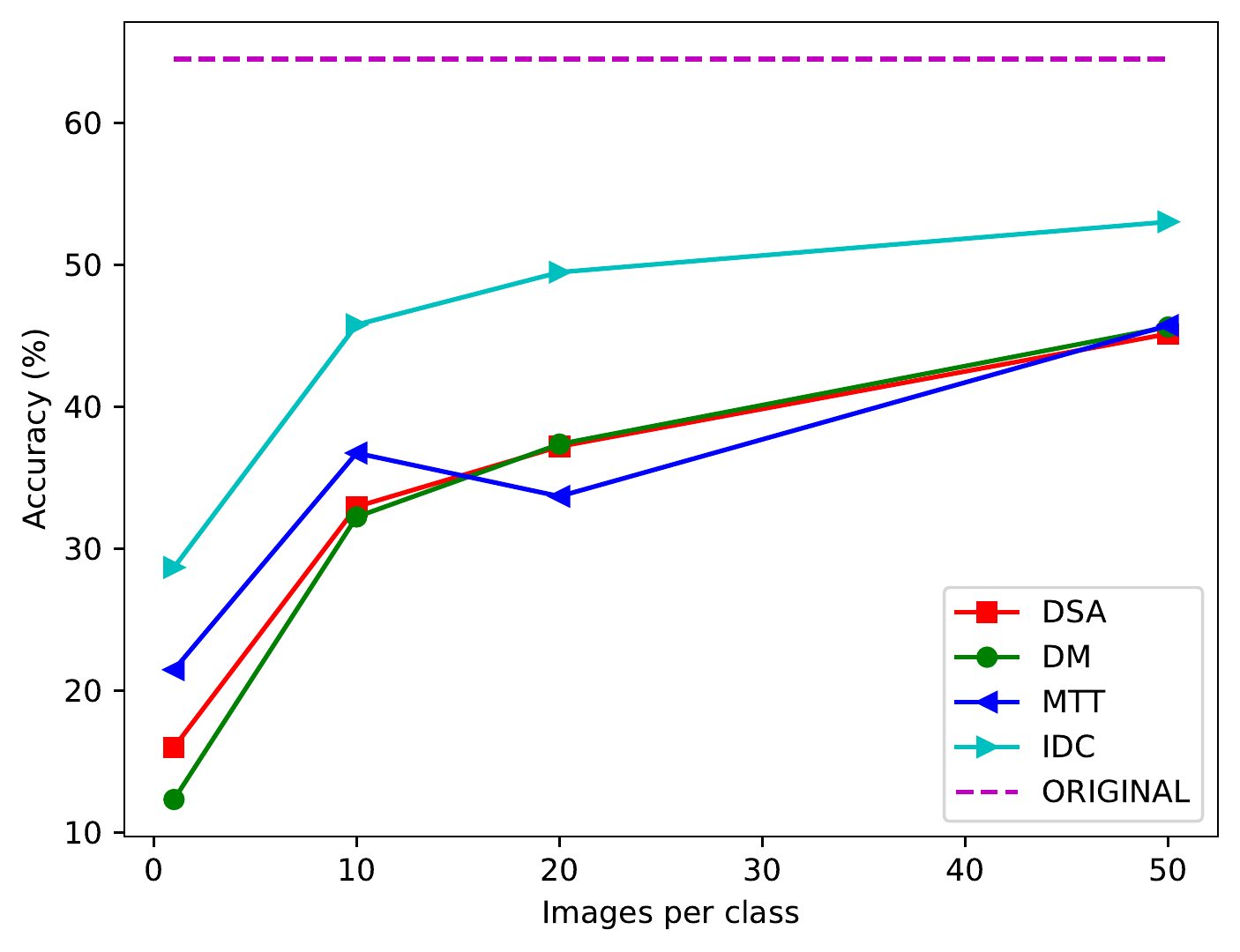}}\hfil 
\subfloat[CIFAR10]{\label{fig:cifar10cross}\includegraphics[width=0.22\textwidth]{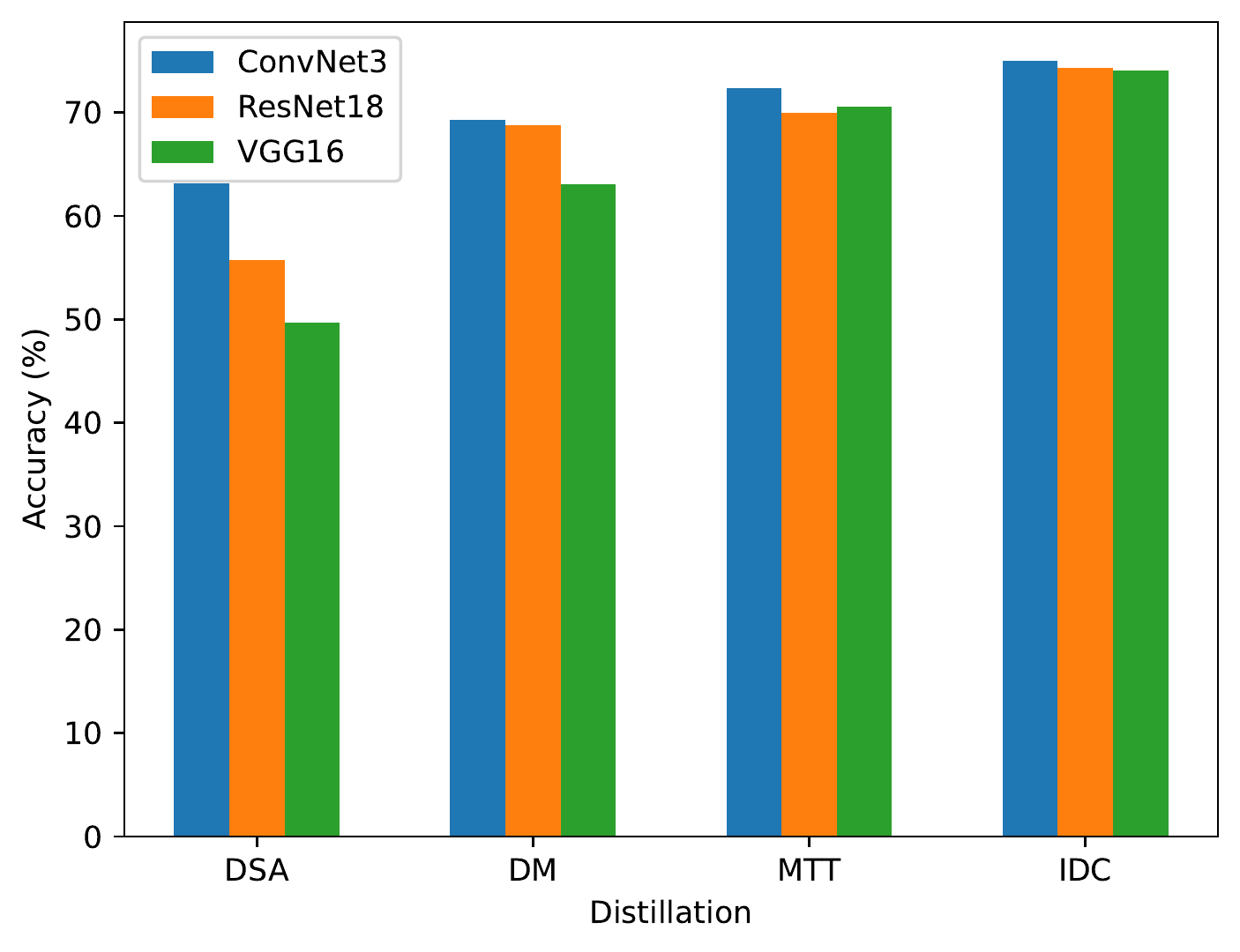}}\hfil   
\subfloat[CIFAR100]{\label{fig:cifar100cross}\includegraphics[width=0.22\textwidth]{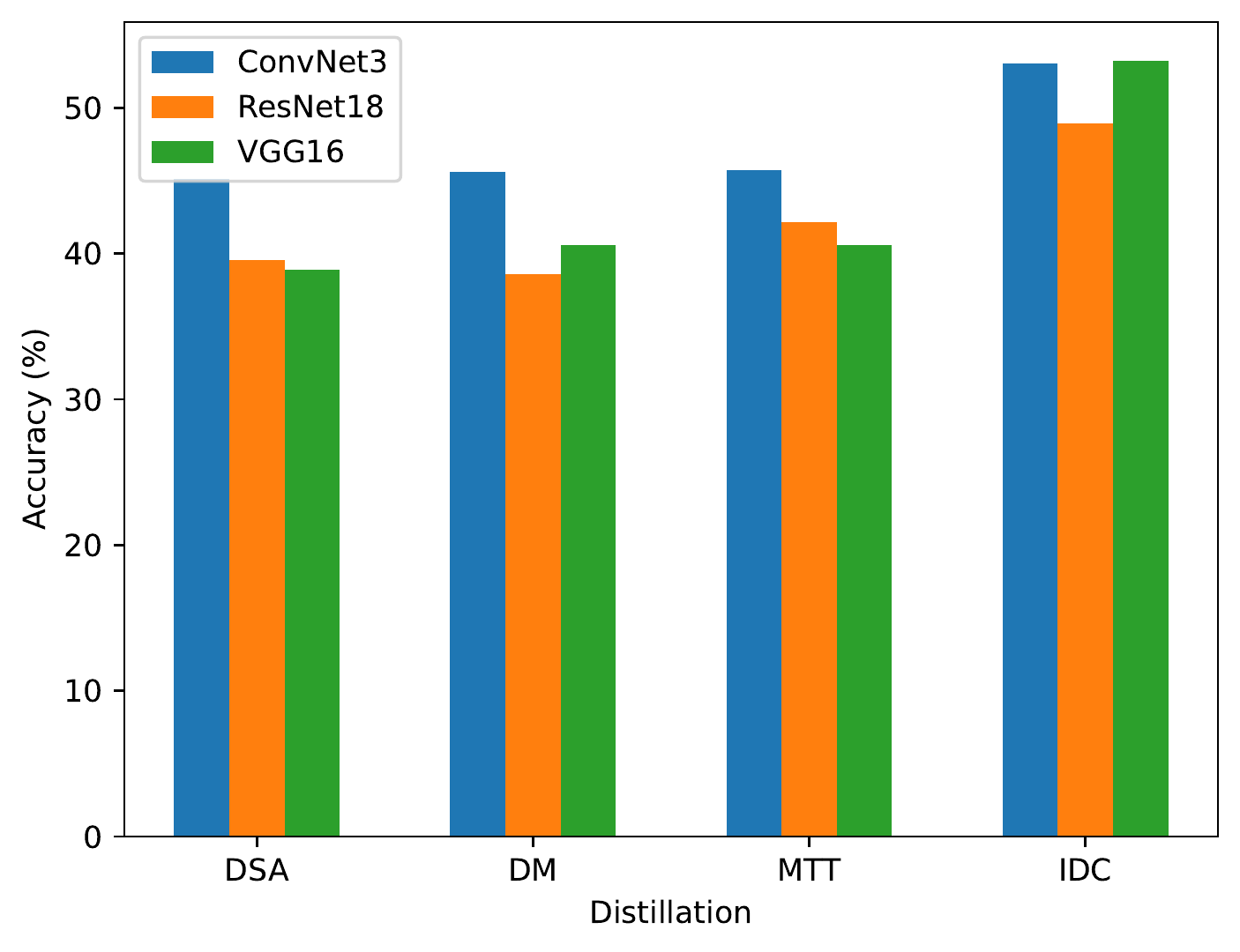}}\hfil
\caption{Test accuracies of models trained on the distilled data}
\label{fig:overall}
\end{figure*}

In this section, We comprehensively evaluate the properties of dataset distillation crafted by the state-of-the-art distillation method through extensive experiments.

We have designed specific experiments for each property and layout the details for the evaluations further. Section ~\ref{sec:overall} introduces the overall performance of the various approaches. 
Readers can grab a quick overview of current developments in this domain. We report the performance of the distilled data under different compression ratio with the  model performance and the generalization capability across various network architectures. In Section \ref{sec:privacy}, \ref{sec:robust} and \ref{sec:fair}, we compare the performance of the models trained on the original and distilled data and analyze the changes that dataset distillation brings to the deep models in terms of privacy, robustness, and fairness.

\subsection{Experimenetal Setup}

\textbf{Performance Evaluation Setup} Using the code released by authors, we run Convnet3 to synthesize 1/10/20/50/100 and 1/10/20/50 images per classes distilled dataset for CIFAR10 and CIFAR100 respectively. 
we use default hyperparameters and training configurations of considered methods, except the following exceptions: 1) For IPC 50 and 100, we manually tune the number of iterations for outer and inner optimization for DSA and DM since they're undefined in the released source code. 2) In MTT, ZCA whitening, a plugable mechanism to improve the performance of distillation dataset as reported by \cite{cazenavette2022dataset}, is disabled when synthesizing distilled dataset for a better comparison with other methods. Since it is applicable to other considered distillation methods and orthogonal to our benchmark settings. 

To evaluate the accuracy performance of each distilled dataset, we repeatedly train three kinds of networks, namely ConvNet3, Resnet18, VGG16, from scratch three times and train them for 2000 epochs to ensure model convergence. The number of epochs we use is different from that of provided by \cite{kim2022dataset} since we found that MTT cannot achieve the optimal accuracy reported in \cite{cazenavette2022dataset} within 300 epochs. For a fair comparison, we train all networks for 2000 epochs instead.

\textbf{Privacy Evaluation Setup} To obtain distilled dataset for performing MIA attacks, we use the distillation algorithm described previously to obtain corresponding synthetic dataset via the same settings we describe above except using a \textit{conceptual} original dataset which is drawn 40, 80, 200, 400 samples per class on CIFAR10 randomly. We don't report the results about MTT since we failed to synthsize the corresponding distilled dataset.

\textbf{Robustness Evaluation Setup} 
We use adversarial examples to attack the target model and measure the robustness of the model through adversarial accuracy. We have considered training the target models to the same accuracy. However, this approach is unfair to those methods that can be trained to very high accuracy. Therefore, we choose to test the adversarial accuracy of the best models obtained by different approaches.

\subsection{Dataset Distillation Performance}\label{sec:overall}

We evaluate the cross architecture performance of different condensation methods on the CIFAR10  datasets with IPC ranging from 1 to 100, and CIFAR100 dataset with IPC ranging from 1 to 50. Figure~\ref{fig:cifar10acc} and Figure~\ref{fig:cifar100acc} illustrate the test accuracy of the models trained on dataset after distillation on CIFAR10 and CIFAR100 respectively. The test accuracy of the model increases as the number of images per class grows. IDC outperforms MTT, DM, and DSA on CIFAR10. On CIFAR100, IDC still performs the best, the performance of DSA and DM are quite close, and MTT is only second to IDC when the number of images per class is small, despite the fluctuations in the curve. Due to space limit, we only show the curves here, the complete tables containing various distillation methods, architectures, and numbers of images per classes can be found in the appendix. 

The learned synthetic images of CIFAR10 are visualized in Figure~\ref{fig:synthetic_dataset}. We found that the images generated by the four methods correspond to four different styles. DM-generated images are difficult to recognize its content even though it is initialized with real images; in contrast, DSA-generated content becomes less noisy. The pictures generated by MTT are more colorful but contain a lot of strange distortions, and the style is closer to comics. IDC is different from others because, through the multi-formation function, it learns to include four down-sampled subimages in one image. These subimages look clearer and closer to the real image. This method is controversial because it provides more valid information.

Based on the visualization result, we believe that IDC's higher accuracy mainly due to its use of data parameterization, where more down-sampled training samples are included in the same image. It actually provides more valid information to the model. Although there is no study that states that the validity of distilled data is related to the identifiability of the images, IDC and MTT show a substantial lead in accuracy and identifiability at CIFAR10, when the number of images per class 10. We have similar findings on the CIFAR100 dataset, which is illustrated in the appendix. 




Figure~\ref{fig:cifar10cross} and Figure~\ref{fig:cifar100cross} demonstrate the generalization capability of different methods on CIFAR10 and CIFAR100. Compared with ConvNet, the test accuracies drop when transferring to ResNet18 and VGG16. 
Moreover, the relative ranking of different distilled dataset might not necessarily be preserved when transferred to different architectures. Our results are consistent with reported in~\cite{zhou2022dataset}.
%



\subsection{Privacy Evaluation}\label{sec:privacy}
In many related works \cite{zhou2022dataset,chen2022private}, dataset distillations have been implemented as a compression technique to protect membership information of training dataset. However, we think their experiments ignore the most fundamental fact, namely the distillation ratio. To verify this, we fist construct training sets of different sizes, and then perform dataset distillation. Specifically, we randomly select 40, 80, 200, 400, 4000 samples from each class in CIFAR10 to form a training set. Besides, we also examined the impact of different initializations on MIA attacks. The results as shown in Table~\ref{tab:miasize}. 
Note that we do not show the results for MTT because we always fail when using MTT for distillation on smaller datasets. 
We apply AUC as a metric to measure how success an MIA attack is. The higher AUC value refers to the higher vulnerability of a model against MIA attacks. Specifically, when the AUC value is equal to 50\%, the model can perfectly prevent MIA attacks. We find that the distillation rate has a positive correlation with the vulnerability of the model against MIA attacks. Specifically, when the distillation ratio is equal to 25\%, the AUC of MIA attacks for IDC can reach an impressive value, 98.07\%. Due to the parameterisation applied in IDC, the distilled datasets contain almost the same information as the original datasets. Furthermore, the experimental results also indicate that distilled datasets initialised with real images contain a greater privacy risk. Additionally, the privacy risk of IDC is significantly higher than the other two methods, which is clearly a side effect of the greater amount of information. 

Besides, we investigated how the number of classes affected MIA attacks. The experiments were conducted on the above-mentioned tailored datasets with a size of 80 and 400, respectively. We changed the number of classes in the classification model from 2 to 5, and 10. The result shown in Table~\ref{tab:miaclass} demonstrate that  even on the distilled dataset, the number of classifications of the target model helps to understand how much privacy information leaked. The larger the number of classes, the more information an attacker can get about the internal state of the model, which is also observed by ~\cite{shokri2017membership}. 

\begin{table}[]
\begin{center}
    
\begin{tabular}{ccccc}
\hline
Origin               & Init   & DSA            & DM            & IDC            \\ \hline
\multirow{2}{*}{40}  & random & 73.67          & 75.08            & 86.47          \\ \cline{2-5} 
                     & real   & \textbf{76.46} & \textbf{77.73}  & \textbf{98.07} \\ \hline
\multirow{2}{*}{80}  & random & \textbf{63.40} & \textbf{63.64}   & 69.86          \\ \cline{2-5} 
                     & real   & 61.89          & 63.23           & \textbf{71.59} \\ \hline
\multirow{2}{*}{200} & random & \textbf{55.06} & 55.15           & 59.09          \\ \cline{2-5} 
                     & real   & 55.03          & \textbf{55.59}    & \textbf{59.73} \\ \hline
\multirow{2}{*}{400} & random & 52.38          & 52.35             & 54.62          \\ \cline{2-5} 
                     & real   & \textbf{52.52} & \textbf{52.89}     & \textbf{56.34} \\ \hline
\end{tabular}

\caption{MIA attack AUC on CIFAR10 with different original sizes and initialization. In both settings, the distilled datasets contain 10 images per class. }
\label{tab:miasize}
\end{center}

\end{table}

\begin{table}[]
\begin{center}
    
\begin{tabular}{lllll}
\hline
Origin               & Classes & DSA            & DM             & IDC            \\ \hline
\multirow{3}{*}{80}  & 2       & 49.77          & 49.78           & 56.80          \\ \cline{2-5} 
                     & 5       & 52.46          & 53.30           & 65.30          \\ \cline{2-5} 
                     & 10      & \textbf{61.89} & \textbf{63.20} & \textbf{71.50} \\ \hline
\multirow{3}{*}{400} & 2       & 50.38          & 50.09            & 52.55          \\ \cline{2-5} 
                     & 5       & 50.46          & 50.67               & 53.24          \\ \cline{2-5} 
                     & 10      & \textbf{52.52} & \textbf{52.89}       & \textbf{54.45} \\ \hline
\end{tabular}
\caption{MIA attack AUC on CIFAR10 with different number of classes.}
\label{tab:miaclass}
\end{center}
\end{table}

\subsection{Fairness Evaluation}\label{sec:fair}
\begin{figure}
\centering
    \subfloat[Normalized accuracy on CIFAR10 at IPC-50]{\label{fig:fairness-classwise-acc-diff-methods} \includegraphics[width=0.5\textwidth]{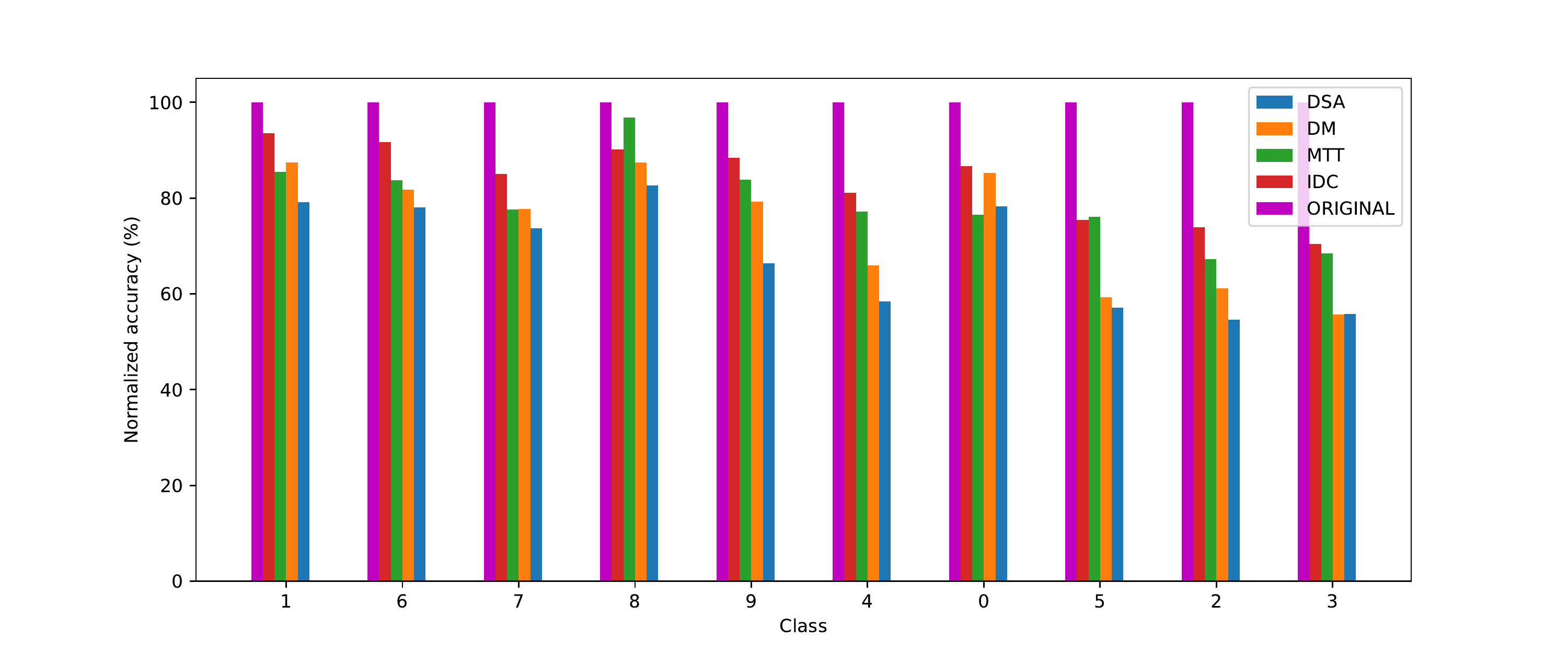}}
    \hfill
    \subfloat[Accuracy on CIFAR10 of original dataset and DSA's dataset at IPC-1]{\label{fig:fairness-classwise-acc-diff-methods-dsa-ipc1} \includegraphics[width=0.5\textwidth]{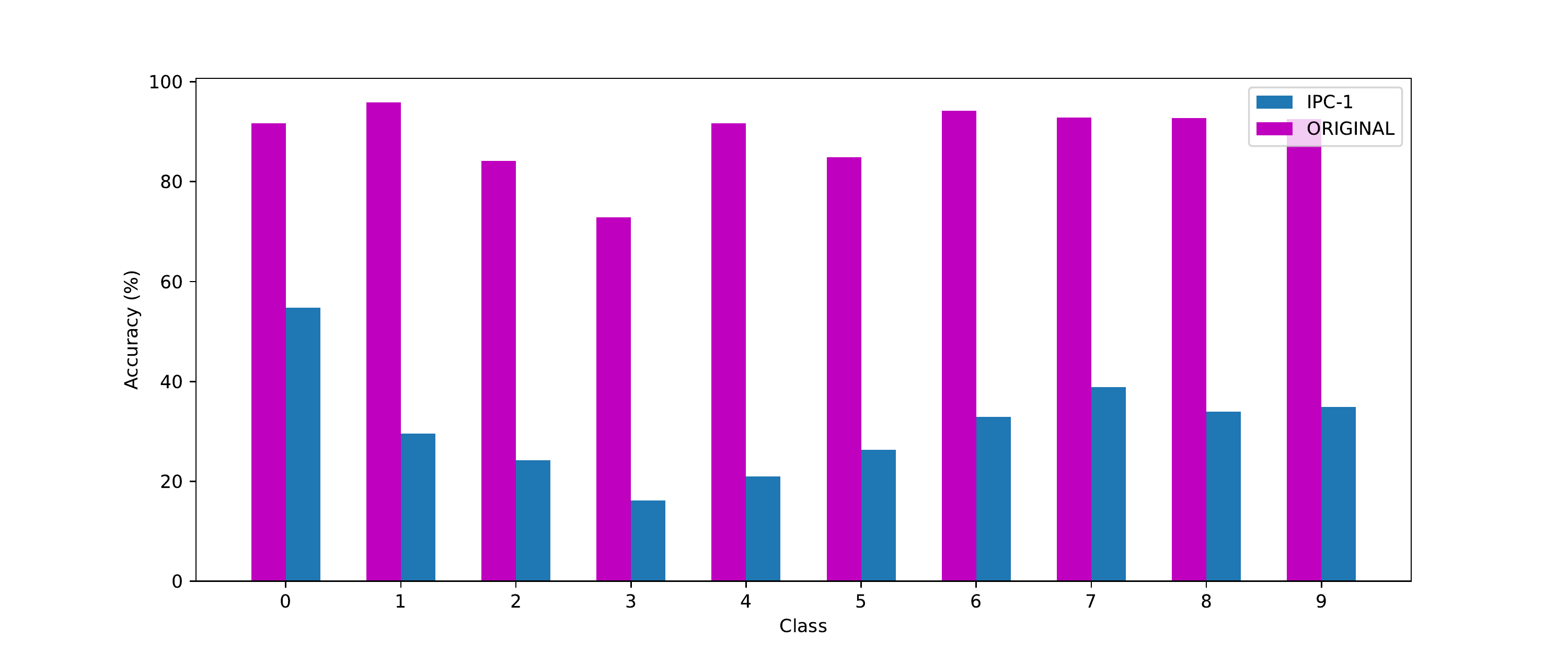}}
    \hfill
    \caption{
    Figure~\ref{fig:fairness-classwise-acc-diff-methods} presented the normalized accuracy of different methods at IPC-50. The order the class is sorted descending according to the original accuracy obtained from original model. 
    Figure~\ref{fig:fairness-classwise-acc-diff-methods-dsa-ipc1} displayed the accuracy of each class evaluated on DSA at IPC-1
    }\label{fig:classwise-acc}
\end{figure}

We first show in Figure~\ref{fig:fairness-classwise-acc-diff-methods} how much accuracy drops on different classes after using dataset distillation. In order to conclude more intuitively, we normalize real accuracy of each class from original dataset to 1. In addtion, we reorder the classes from high to low according to the real accuracy, i.e., the original model performs best on Class 1 and worst on 3. In this way, the variation of the same method on different classes represents the unfairness brought by the distillation. Secondly, we find that different classes have different difficulties in learning through distillation. For example, in classes with relatively higher real accuracy, such as Class 1, 6, 7, and 8, the normalized accuracy obtained by different methods is also relative higher. For Class 2 and 3, the normalized accuracy obtained by different methods is relatively low, which means that more information is lost in the distillation process and learning is more difficult. It can be seen that the loss of distillation information or learning difficulty is closely related to the performance of the original model on these categories. This conclusion inspires that we may need to increase IPC for classes with poor real performance to retain more training information.

\begin{figure}[h]
\centering
\subfloat[Loss-based fairness]{\label{fig:fairness-loss} \includegraphics[scale=0.25]{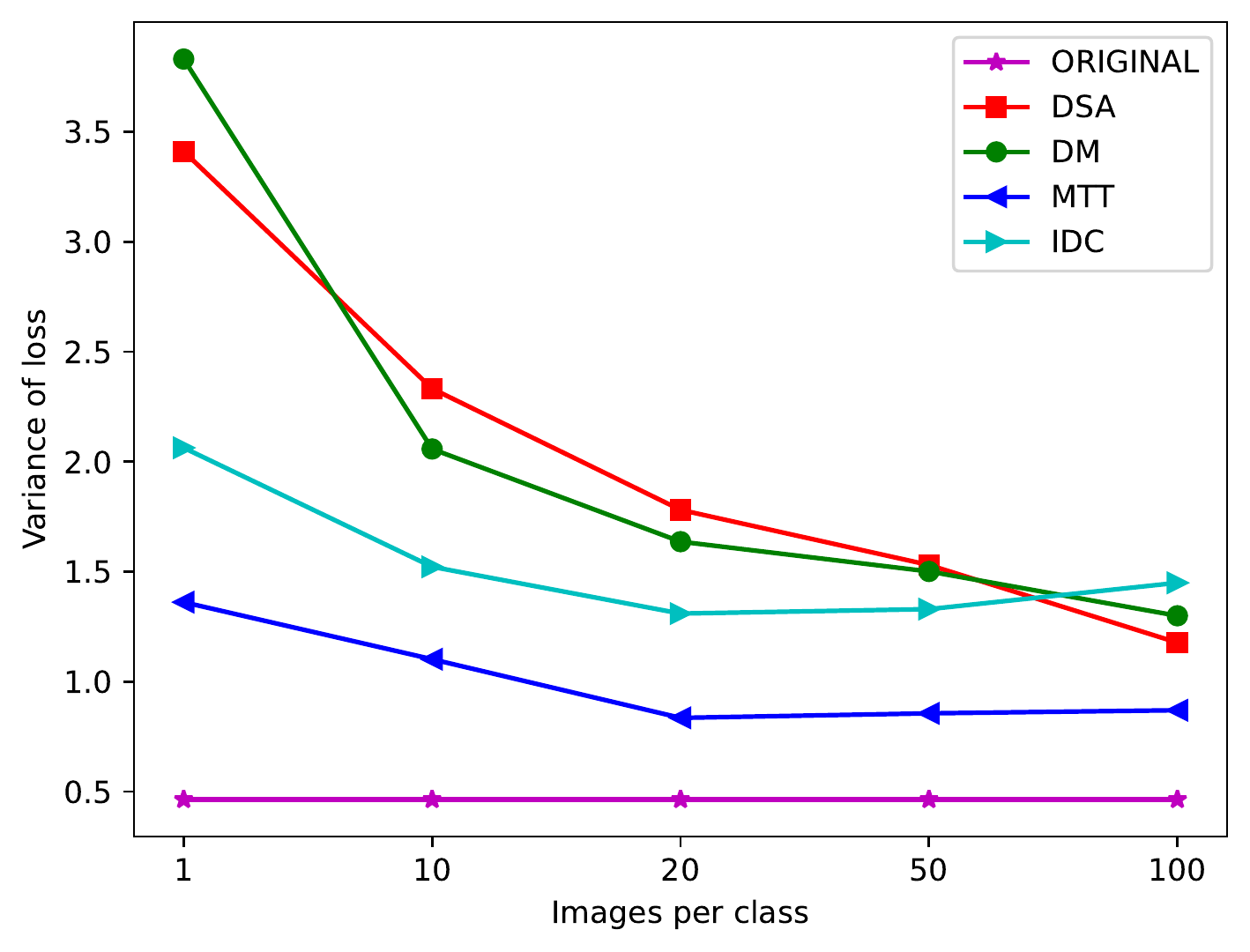}}
\hfill
\subfloat[Class-wise accuracy-based fairness]{\label{fig:fairness-acc} \includegraphics[scale=0.25]{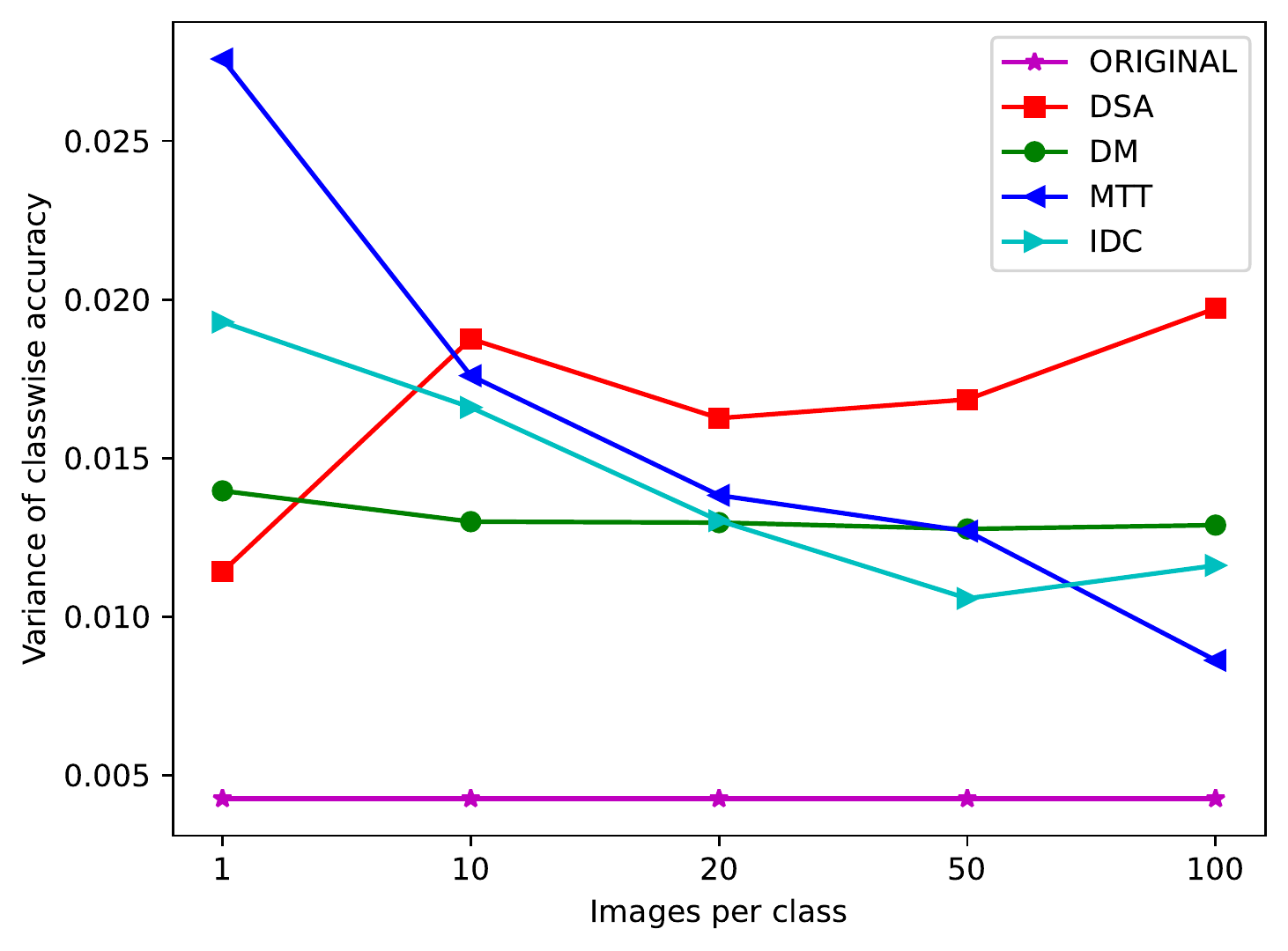}}
\caption{Fairness of a model evaluated according to Definition~\ref{def:performance-fairness}}
\label{fig:fairness}
\end{figure}

We then plot the variance across different classes in terms of loss and accuracy in Figure~\ref{fig:fairness-loss} and Figure~\ref{fig:fairness-acc} based on Definition~\ref{def:performance-fairness}. We clearly find that as the IPC increases, the model's fairness also improve since the variance of loss decreases. We also find different methods have different intrinsic fairness properties. Overall, the MTT consistently outperforms the IDC, DM and DSA in terms of loss-based fairness. As to accuracy-based fairness, we find that the situation is somewhat complicated, with DSA performing best when IPC equals 1. For this reason, we plot accuracy of original and DSA with IPC=1 in Figure~\ref{fig:fairness-classwise-acc-diff-methods-dsa-ipc1}. We find that this may be due to the fact that the model performs poorly on different categories. To further explain the unfairness between the categories, we visualize the distribution of features across the synthetic data when IPC=50 using t-SNE in Figure~\ref{fig:tsne}. We find that MTT have the best visualisation result, with almost all distillation features falling in the corresponding cluster, while the other three methods all have distillation feature falling in other clusters. In addition, there is differences across categories. For example, in Figure~\ref{fig:tsne-dm}, the features corresponding to Class 0 and Class 2 both have some distillation features falling in each other's cluster, but almost none fall into Class 1. This could be a explanation of unfairness. We suggest using a loss-based fairness rather than accuracy, as \textit{softmax} provides more information across classes, whereas accuracy loses such information due to \text{argmax} operation.
%



\subsection{Robustness Evaluation}\label{sec:robust}

The robustness accuracy of different methods on CIFAR10 and CIFAR100 are illustrated in  Figure~\ref{fig:all-robustallcifar10} and Figure~\ref{fig:all-robustallcifar100}, where the horizontal coordinate represents the strength of the added noise, the vertical coordinate represents the accuracy in the classification task, and higher values represent greater resistance to robustness attacks. The first picture in each row shows the results of different distillation methods when the number of images per class (IPC) is set to 50, and the following four  represent the change of robustness with IPC in different methods.
We found that three of the four methods we experimented with: DM, DSA, and MTT were all able to outperform the original model in terms of robustness when the noise exceeded a certain intensity. While on CIFAR100, only DSA and DM would perform better when the noise was sufficiently high. In both datasets, the MTT method always exhibited the worst robustness. We also compare the accuracies with varying IPCs, however,We don't not find a strong relationship between robustness and IPC.

\section{Related Work}
\label{sec:related}

Wang et al.~\cite{wang2018dataset} was the first to propose the concept of dataset distillation (also known as dataset condensation). They proposed to using bi-level optimization to synthesize a compact dataset that has better storage efficiency than the coresets. Since then, more and more methods have been proposed to improve distillation performance, for example using distribution matching~\cite{Zhao2021DatasetCW}, gradient matching~\cite{zhao2021datasetdc}, training trajectory matching~\cite{cazenavette2022dataset} and augmentation techniques \cite{zhao2021datasetdsa}, kernel methods~\cite{nguyen2020dataset,zhou2022dataset,loo2022efficient}, and data parametrization~\cite{kim2022dataset} with GAN~\cite{such2020generative,zhao2022synthesizing} or factorization~\cite{lee2022dataset,liu2022dataset}. 

There are other works exploring dataset distillation as in our work. In ~\cite{wang2018dataset}, it was shown that using dataset distillation can make data poisoning attacks more effective. In ~\cite{zhou2022dataset}, it was claimed that dataset distillation is robust to membership inference attacks. Our experimental results empirically show that their conclusions are not entirely correct. In ~\cite{nguyen2020dataset,loo2022efficient}, they built a $\rho-$corrupted dataset to protect privacy, and the image information has a $\rho$ ratio of pixels that remains unchanged during the optimization process. The last one that is similar to our work is DC-Bench~\cite{cui2022dc}. However, they only compared and analyzed existing dataset distillation methods in terms of compression ratio, transferability and neural architecture search. In contrast, we focus on the ML security and potential risks of the dataset distillation.

\section{Conclusion}
\label{sec:conclusion}


\paragraph{Conclusion.} This work systematically evaluates current dataset distillation methods from the perspective of ML security. Specifically, it provides a large-scale benchmark for evaluating dataset distillation, where we design and implement experiments regarding privacy, robustness, and fairness. We find that dataset distillation cannot help prevent membership inference attacks, and it increases the unfairness of model performance across different classes. We also demonstrate experimentally that it impacts the robustness of models to varying degrees. This is the first work that bridges the gap between dataset distillation and security studies. 


\appendix
\section{Hyperparameters \& Hardwares}
For all distilled methods considered in this paper, we use the default hyperparameters given by the authors.
That is, we run 1K iterations to create a synthetic dataset in DSA. As for IPC 50 and 100, we manually set the hyperparameters \textit{inner} and \textit{outer loop} in DSA.
For MTT, we use the same hyperparameters reported by the authors except for disabling ZCA whitening for a fair comparison. For both DM and IDC, the settings are the same as those provided by the authors. Furthermore, we run all our experiments on the same host with Tesla V100 GPU with 32 GB.  As a performance comparison, it takes 2000 epochs to train ConvNet3, Resnet18, and VGG16 to ensure coverage. We also repeat the training 3 times to measure the variance of accuracy. 

\section{Detailed accuracy under different IPCs and models}
In Table~\ref{tab:app-overall}, we report the detailed accuracy of model trained by distilled dataset.
It contains accuracy results about dataset CIFAR10 and CIFAR100 with different IPCs on different models, i.e., ConvNet3, Restnet18 and VGG16.

\begin{table*}[ht]
\begin{center}
\begin{tabular}{|c|c|c|ccccc|c|}
\hline 
Dataset & Image per class & Test Model &  DSA  &  DM   &  MTT  & IDC  & Full dataset \\ \hline \hline
\hline
\multirow{15}{*}{CIFAR10} & \multirow{3}{*}{1}
  & ConvNet3   & 29.50 $\pm$1.22 & 28.84 $\pm$ 0.52 & 41.13 $\pm$0.50 & 48.07 $\pm$ 0.16 & 89.61 $\pm$ 0.26 \\ 
& & Resnet18  & 28.52 $\pm$ 0.29 & 21.74 $\pm$ 0.92 &  34.00 $\pm$ 1.17 & 41.34 $\pm$ 0.54 & 88.85 $\pm$ 5.27 \\ 
& & VGG16     & 20.56 $\pm$ 4.28 & 14.01 $\pm$ 4.01 &  24.15 $\pm$ 1.36    &  37.82 $\pm$ 4.64 & 94.76 $\pm$ 0.04 \\
\cline{2-8}

& \multirow{3}{*}{10}
& ConvNet   & 50.29 $\pm$ 0.23 & 53.07 $\pm$ 0.91 & 63.23 $\pm$0.55 &65.13 $\pm$ 0.12& 89.61 $\pm$ 0.26 \\ 
&  & Resnet18  & 40.26 $\pm$ 0.64 & 38.52 $\pm$ 1.12 & 48.75 $\pm$1.79  & 63.56 $\pm$ 0.91 & 88.85 $\pm$ 5.27\\ 
&  & VGG16     & 36.05 $\pm$ 1.85 & 35.47 $\pm$ 2.29 &  52.05 $\pm$ 0.57   & 59.35 $\pm$ 1.32 & 94.76 $\pm$ 0.04 \\
\cline{2-8}

& \multirow{3}{*}{20}
  & ConvNet   & 57.46 $\pm$0.32 & 60.83 $\pm$0.24 &  65.31 $\pm$ 0.43  & 71.68 $\pm$0.24 & 89.61 $\pm$ 0.26\\ 
  & & Resnet18  & 46.40 $\pm$ 0.39 & 47.41 $\pm$ 0.31  &   61.12 $\pm$ 0.71 & 69.35 $\pm$ 1.18 & 88.85 $\pm$ 5.27 \\ 
&   & VGG16     & 38.10 $\pm$ 0.64 & 48.65 $\pm$ 1.59 &  56.23 $\pm$ 3.15    & 66.81 $\pm$ 0.41 & 94.76 $\pm$ 0.04 \\
\cline{2-8}

& \multirow{3}{*}{50}
& ConvNet   & 61.28 $\pm$ 0.37 & 65.63 $\pm$ 0.08 & 71.64 $\pm$ 0.30 & 75.18 $\pm$ 0.40 & 89.61 $\pm$ 0.26 \\ 
& & Resnet18  & 52.99 $\pm$ 0.18 & 62.34 $\pm$ 0.57 &   68.94 $\pm$ 0.14  & 72.14 $\pm$ 0.10 &88.85 $\pm$ 5.27\\ 
& & VGG16     & 45.00 $\pm$ 0.27 & 56.81 $\pm$ 3.89 & 69.08 $\pm$ 0.22 & 72.95 $\pm$ 0.26 & 94.76 $\pm$ 0.04 \\ 
\cline{2-8}

& \multirow{3}{*}{100}
 & ConvNet   & 63.14 $\pm$ 0.14 & 69.31 $\pm$ 0.16 &  72.39 $\pm$ 0.13     & 75.00 $\pm$ 0.04 & 89.61 $\pm$ 0.26  \\ 
 & & Resnet18  & 55.71 $\pm$ 0.08 & 68.81 $\pm$ 0.49 &   69.94 $\pm$ 0.21   & 74.32 $\pm$ 0.26 & 88.85 $\pm$ 5.27 \\ 
&   & VGG16     & 49.66 $\pm$ 0.85 & 63.07 $\pm$ 4.08 &    70.60 $\pm$  0.27 & 74.02 $\pm$ 0.02 & 94.76 $\pm$ 0.04 \\ 
\cline{1-8}

\multirow{15}{*}{CIFAR100} & \multirow{3}{*}{1}
  & ConvNet   & 16.01 $\pm$ 0.26 & 12.33 $\pm$ 0.11 & 21.47 $\pm$ 0.36 & 28.68 $\pm$ 0.41 & 64.5 $\pm$  0.15\\
&  & Resnet18  & 10.1 $\pm$ 0.51 & 3.02$\pm$ 0.14 & 10.51 $\pm$ 0.03 & 21.67 $\pm$ 0.15 & 72.57 $\pm$ 0.14  \\ 
& & VGG16     &  10.67 $\pm$ 0.35 & 5.06 $\pm$ 0.43 &  7.33 $\pm$ 0.19 & 21.43 $\pm$ 0.97 & 74.05 $\pm$ 4.54\\
\cline{2-8}

& \multirow{3}{*}{10}
& ConvNet   &  32.96 $\pm$ 0.19 & 32.24 $\pm$ 0.26 & 36.74 $\pm$ 0.17 & 45.78 $\pm$ 0.24 & 64.35 $\pm$  0.15\\ 
& & Resnet18  & 24.14 $\pm$0.27 & 26.45 $\pm$ 0.67 &  31.70 $\pm$ 0.21 & 41.42 $\pm$ 0.40 & 72.71 $\pm$ 0.14\\ 
& & VGG16     & 16.06 $\pm$ 0.86 & 24.63 $\pm$ 2.28  & 31.75 $\pm$ 0.61 & 41.86 $\pm$ 0.56 & 74.05 $\pm$ 4.54 \\
\cline{2-8}

& \multirow{3}{*}{20}
      & ConvNet   & 37.20 $\pm$ 0.29  & 37.37 $\pm$0.18 & 33.69 $\pm$ 0.32 & 49.48 $\pm$ 0.03 & 64.35$\pm$  0.15 \\ 
&     & Resnet18  & 28.28 $\pm$0.63 & 33.28 $\pm$ 0.38 & 31.13 $\pm$ 0.42 & 47.24 $\pm$ 0.30 & 72.71  $\pm$ 0.14\\ 
&     & VGG16     & 24.56 $\pm$ 0.48 & 32.83 $\pm$ 0.39 & 27.60  $\pm$ 0.31   & 44.85 $\pm$ 0.11 & 74.05 $\pm$ 4.54 \\
\cline{2-8}

& \multirow{3}{*}{50}
 & ConvNet   &  45.14 $\pm$ 0.22 & 45.62 $\pm$ 0.15 & 45.73 $\pm$ 0.40 & 53.62 $\pm$ 0.20 & 64.35 $\pm$  0.15\\ 
& & Resnet18  & 44.13 $\pm$ 0.21 & 44.65 $\pm$ 0.62 &  42.19 $\pm$ 0.29 & 51.73 $\pm$0.23 & 72.71 $\pm$ 0.14\\ 
& & VGG16     & 38.88 $\pm$ 0.29 &  40.60 $\pm$ 0.42 &  46.46 $\pm$ 0.15 &  53.49 $\pm$ 0.46 &  74.05 $\pm$ 4.54 \\ 
\cline{1-8}





\end{tabular}
\end{center}
\caption{Cross-architecture accuracy results of method DSA, DM, MTT and IDC on CIFAR10 with IPC 1, 10, 20, 50 and 100, and CIFAR100 with IPC 1, 10, 20, 50.}
\label{tab:app-overall}
\end{table*}

\section{Robustness}



\begin{figure*}[t]
    \centering
    \includegraphics[width=0.33\textwidth]{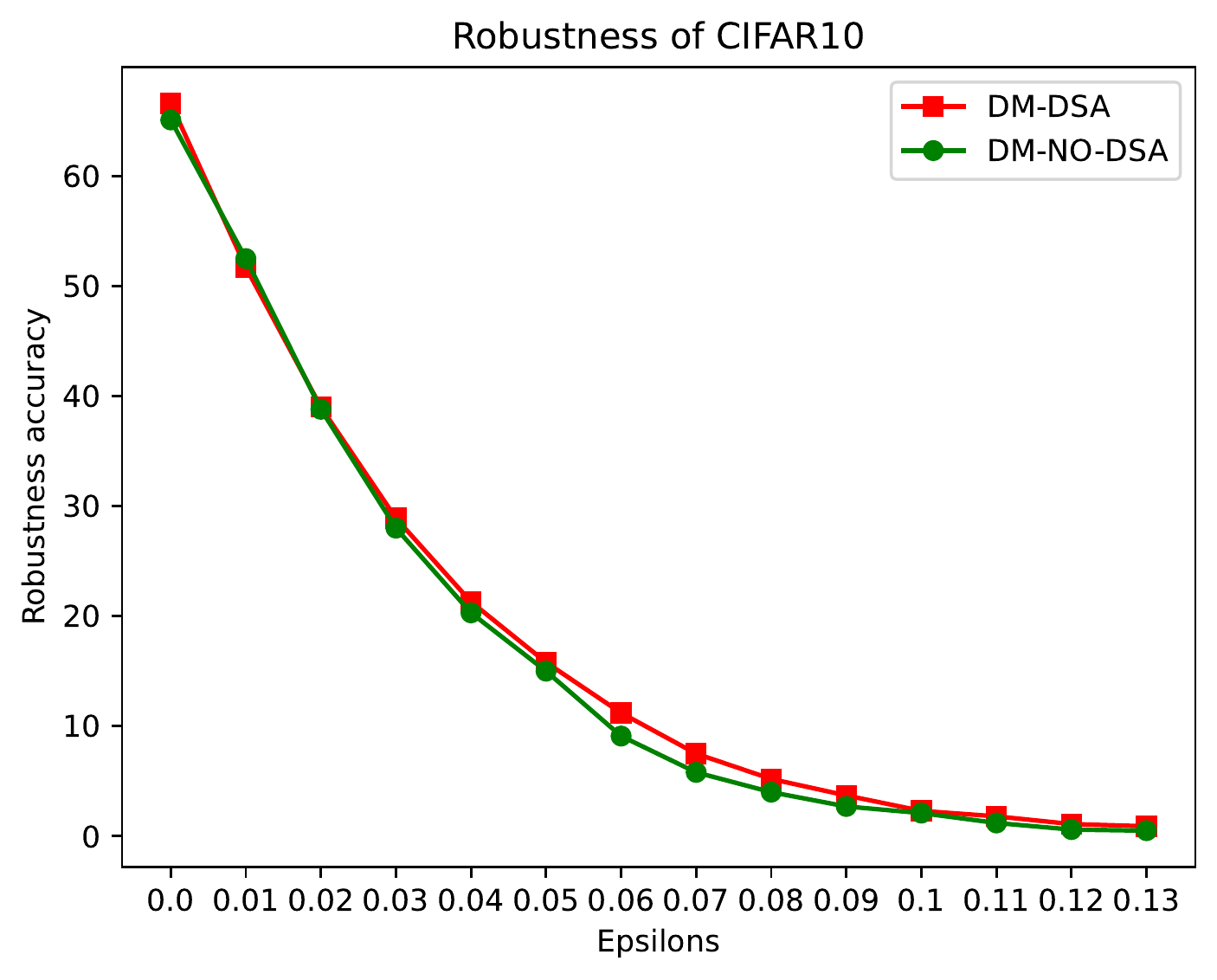}
    \includegraphics[width=0.33\textwidth]{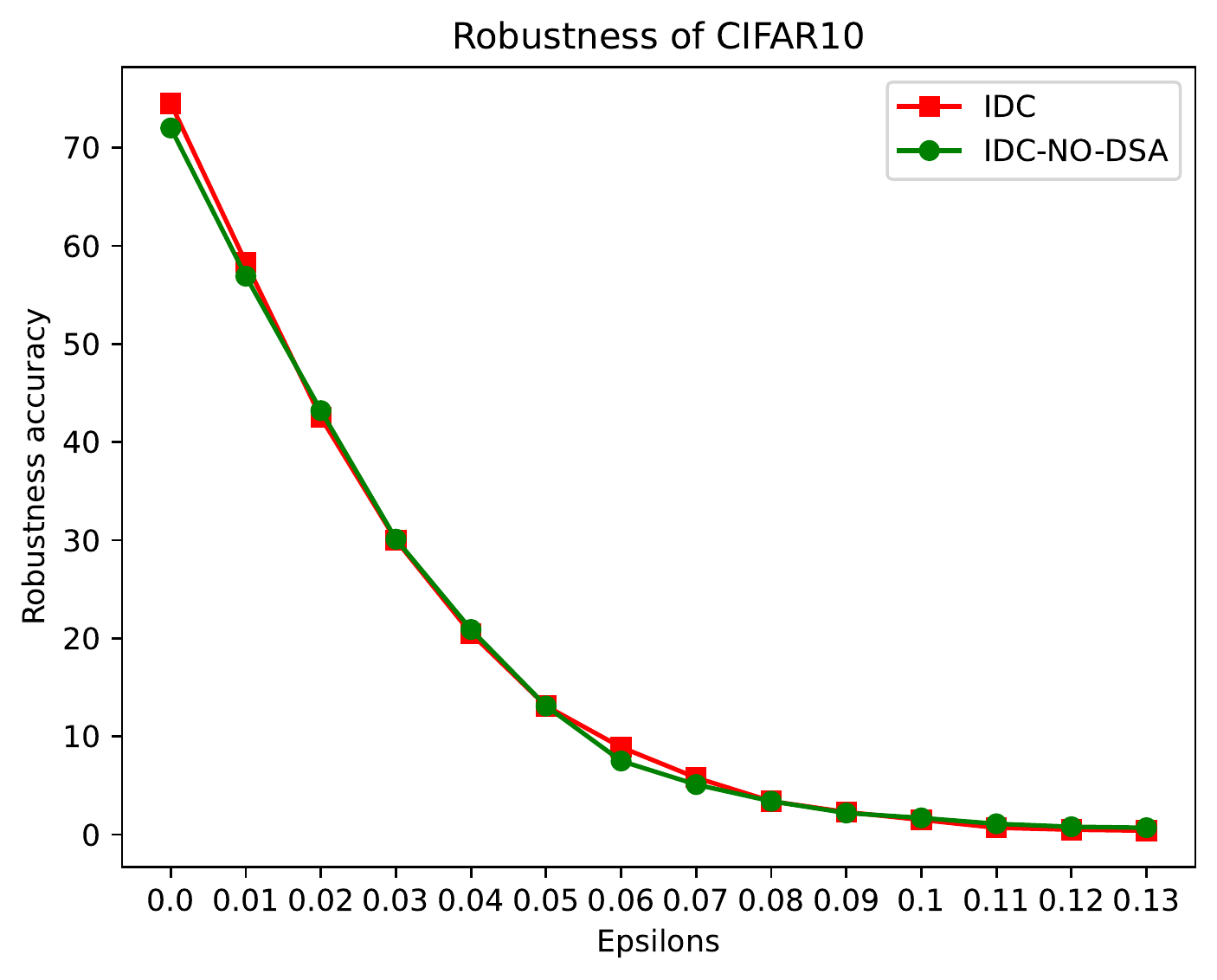}
        \includegraphics[width=0.33\textwidth]{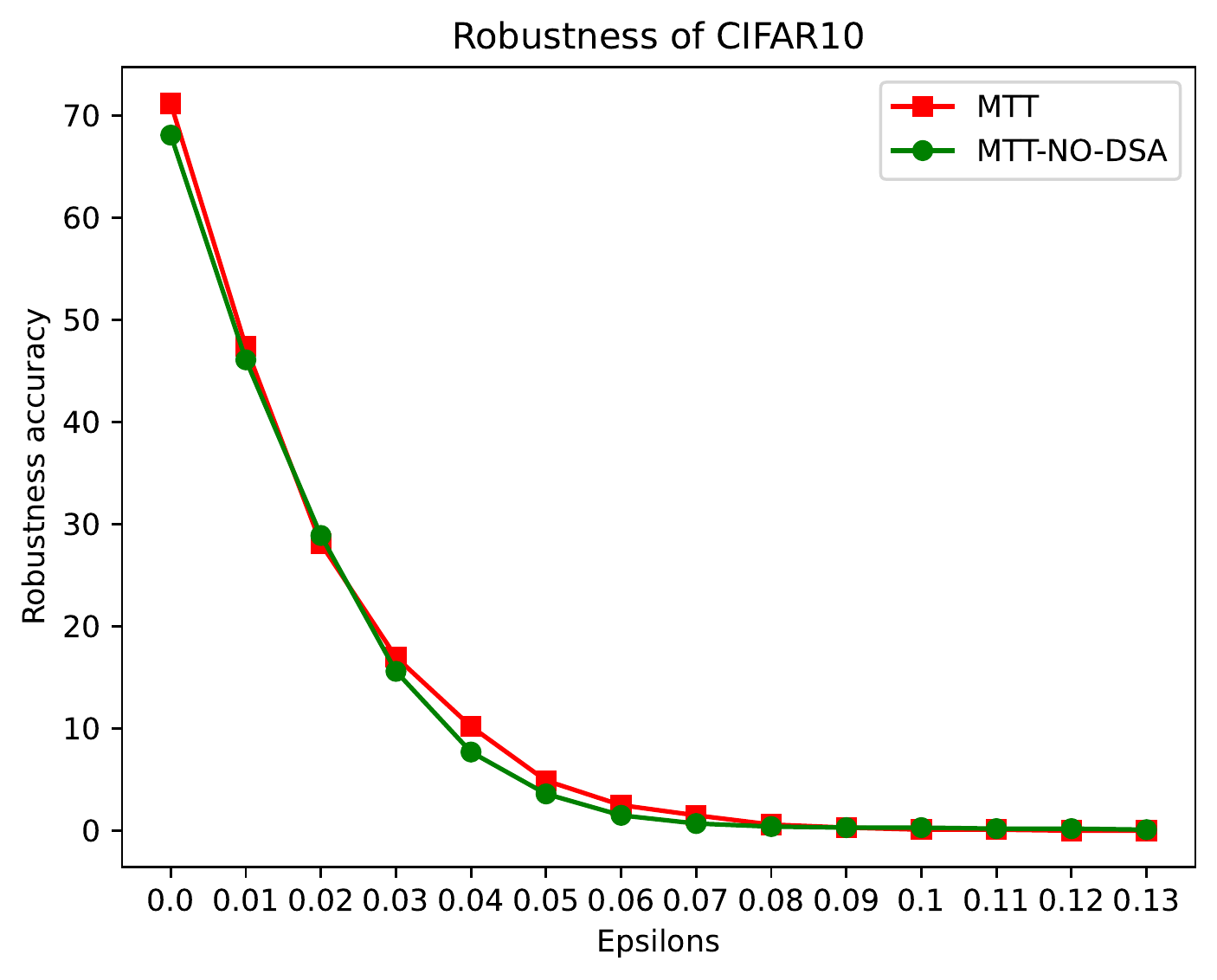}
    \caption{Robust accuracy of dataset trained by enabling DSA and disabling DSA}
    \label{fig:app-robustness-no-dsa}
\end{figure*}

In Figure~\ref{fig:app-robustness-no-dsa}, we investigate the impact of DSA augmentation on DM, MTT, and IDC methods. We generate two different distilled datasets by enabling and disabling DSA augmentation of each algorithm. Then, we train the models on each dataset to obtain the best validation accuracy. 
Last, we use DeepFoolAttack to evaluate the robustness of each model to analyze the influence of DSA. We see that DSA augmentation can improve the robustness to some extent since the red line in Figure~\ref{fig:app-robustness-no-dsa} is always slightly above the green curve. 

\section{Visualization of distilled dataset}
\begin{figure*}[t]
    \centering
    \subfloat[DSA]{\label{fig:app-distilled-cifar100-dsa} \includegraphics[width=0.5\textwidth]{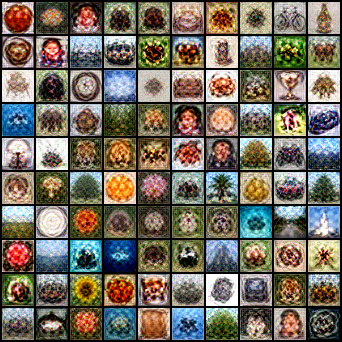}}%
    \subfloat[DM]{\label{fig:app-distilled-cifar100--dm} \includegraphics[width=0.5\textwidth]{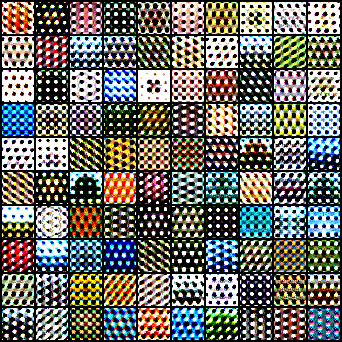}}%
    \hfill
    \subfloat[MTT]{\label{fig:app-distilled-cifar100--mtt} \includegraphics[width=0.5\textwidth]{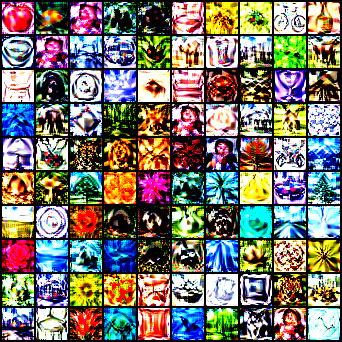}}%
    \subfloat[IDC]{\label{fig:app-distilled-cifar100--idc} \includegraphics[width=0.5\textwidth]{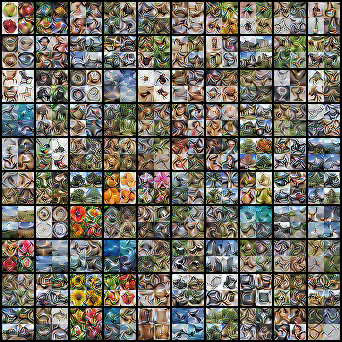}}%
    \hfill
    \caption{Distilled dataset on CIFAR100 with IPC-1}    
    \label{fig:app-cifar100-ipc-1}
\end{figure*}

\begin{figure*}[t]
    \centering
    \subfloat[DSA]{\label{fig:app-distilled-cifar10-dsa} \includegraphics[width=0.45\textwidth]{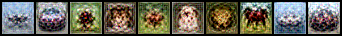}}%
    \subfloat[DM]{\label{fig:app-distilled-cifar10-dm} \includegraphics[width=0.45\textwidth]{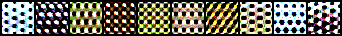}}%
    \hfill
    \subfloat[MTT]{\label{fig:app-distilled-cifar10-mtt} \includegraphics[width=0.45\textwidth]{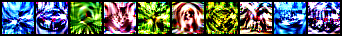}}%
    \subfloat[IDC]{\label{fig:app-distilled-cifar10-idc} \includegraphics[width=0.45\textwidth]{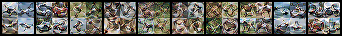}}%
        \hfill
    \centering
    \caption{Distilled dataset on CIFAR10 with IPC-1}
    \label{fig:app-cifar10-ipc-1}
\end{figure*}

In Figure~\ref{fig:app-cifar10-ipc-1}, \ref{fig:app-cifar100-ipc-1}, we show the images of a synthetic dataset of CIFAR10 and CIFAR100 with IPC 1 respectively, which can help us have a better understanding about the dataset pattern of different distilled algorithms. 
Interestingly, we find that the distilled images via DM and MTT can not be recognized directly compared with other algorithms in both CIFAR10 and CIFAR100 datasets, e.g. Figure~\ref{fig:app-distilled-cifar10-dm} looks like a chessboard.  

\section{Distribution of distilled dataset}
\begin{figure*}[h]
\centering
\subfloat[DSA]{\label{fig:app-tsne-dsa-10class} \includegraphics[width=0.5\textwidth]{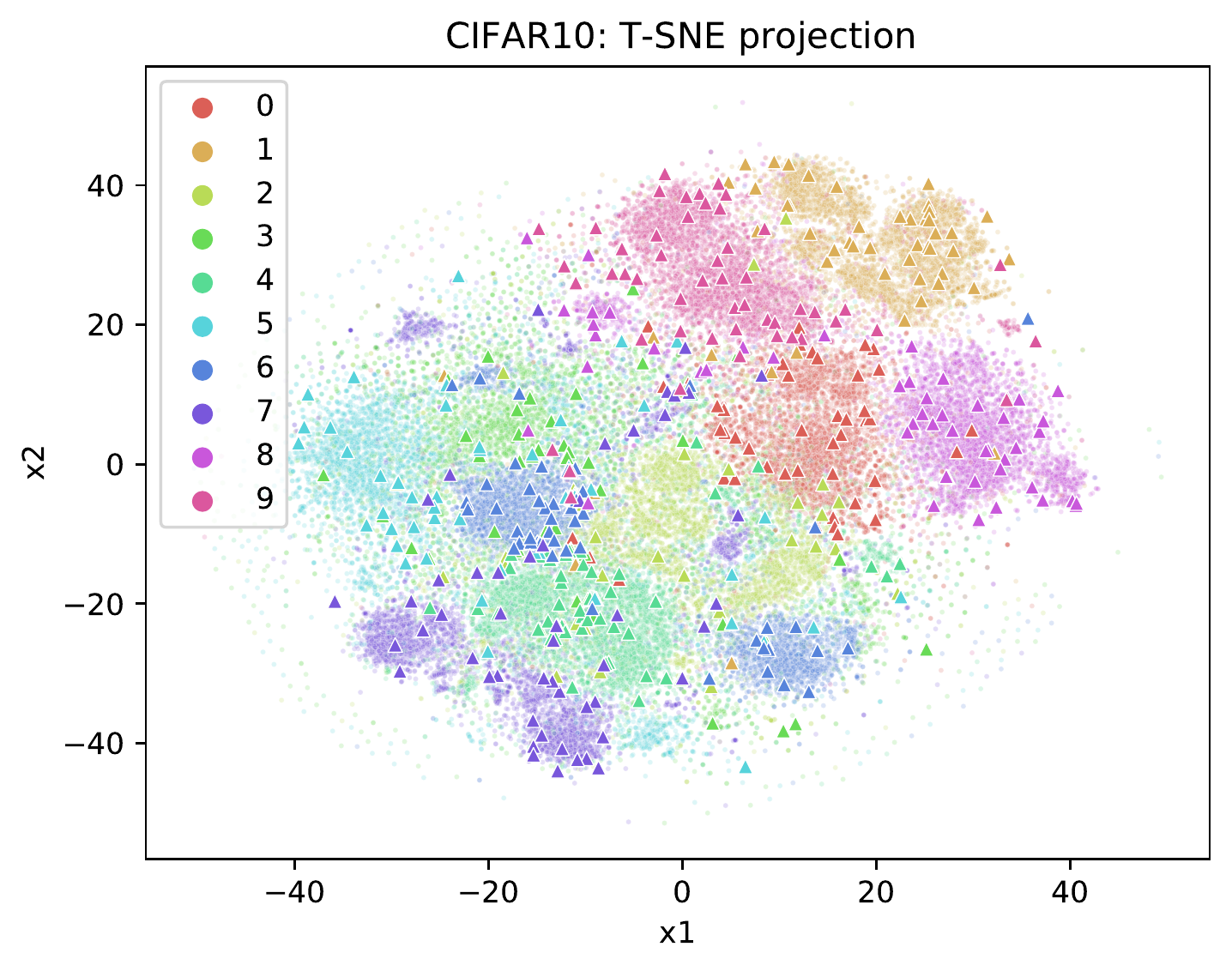}}%
\subfloat[DM]{\label{fig:app-tsne-dm-10class} \includegraphics[width=0.5\textwidth]{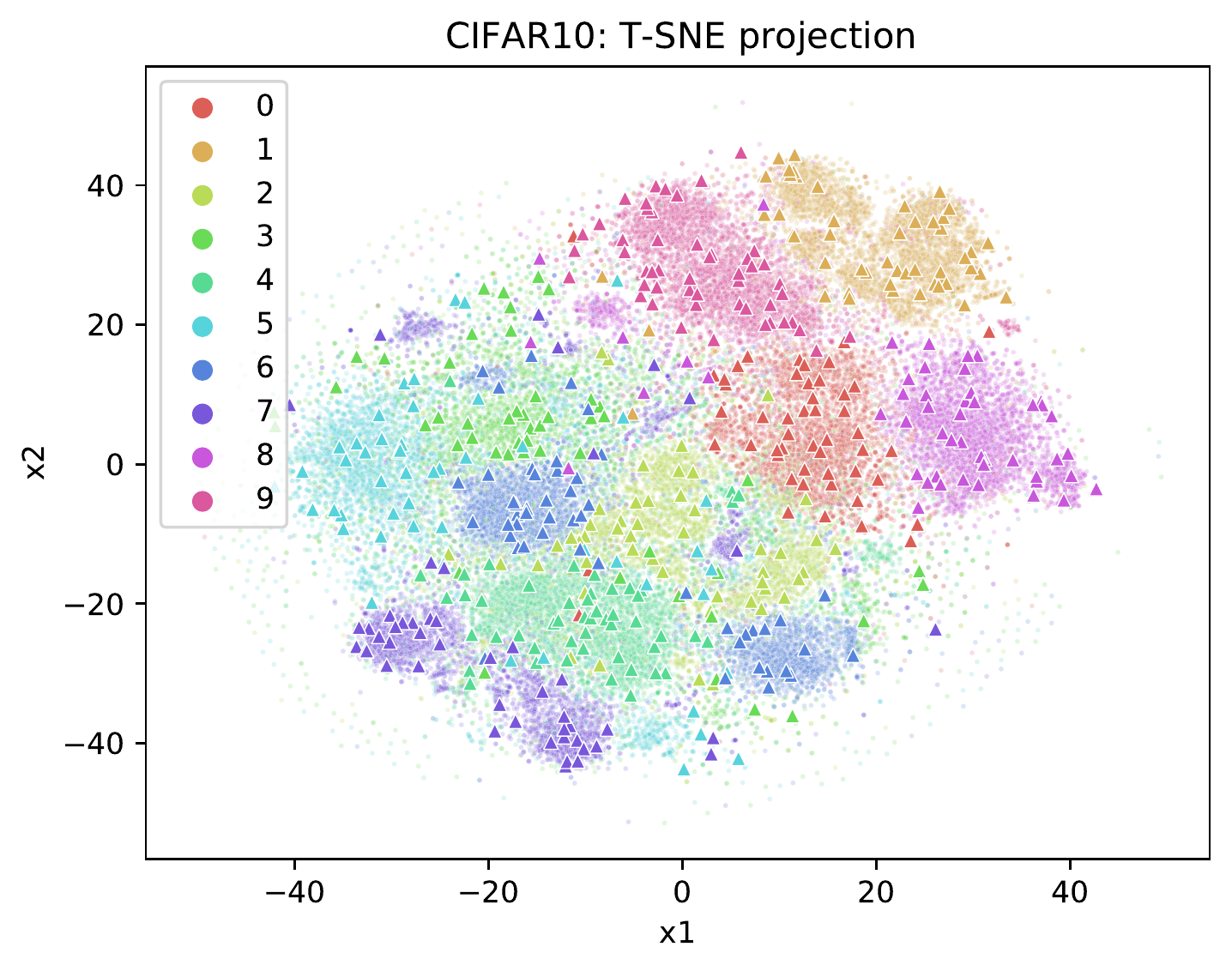}}%
\hfill
\subfloat[MTT]{\label{fig:app-tsne-mtt-10class} \includegraphics[width=0.5\textwidth]{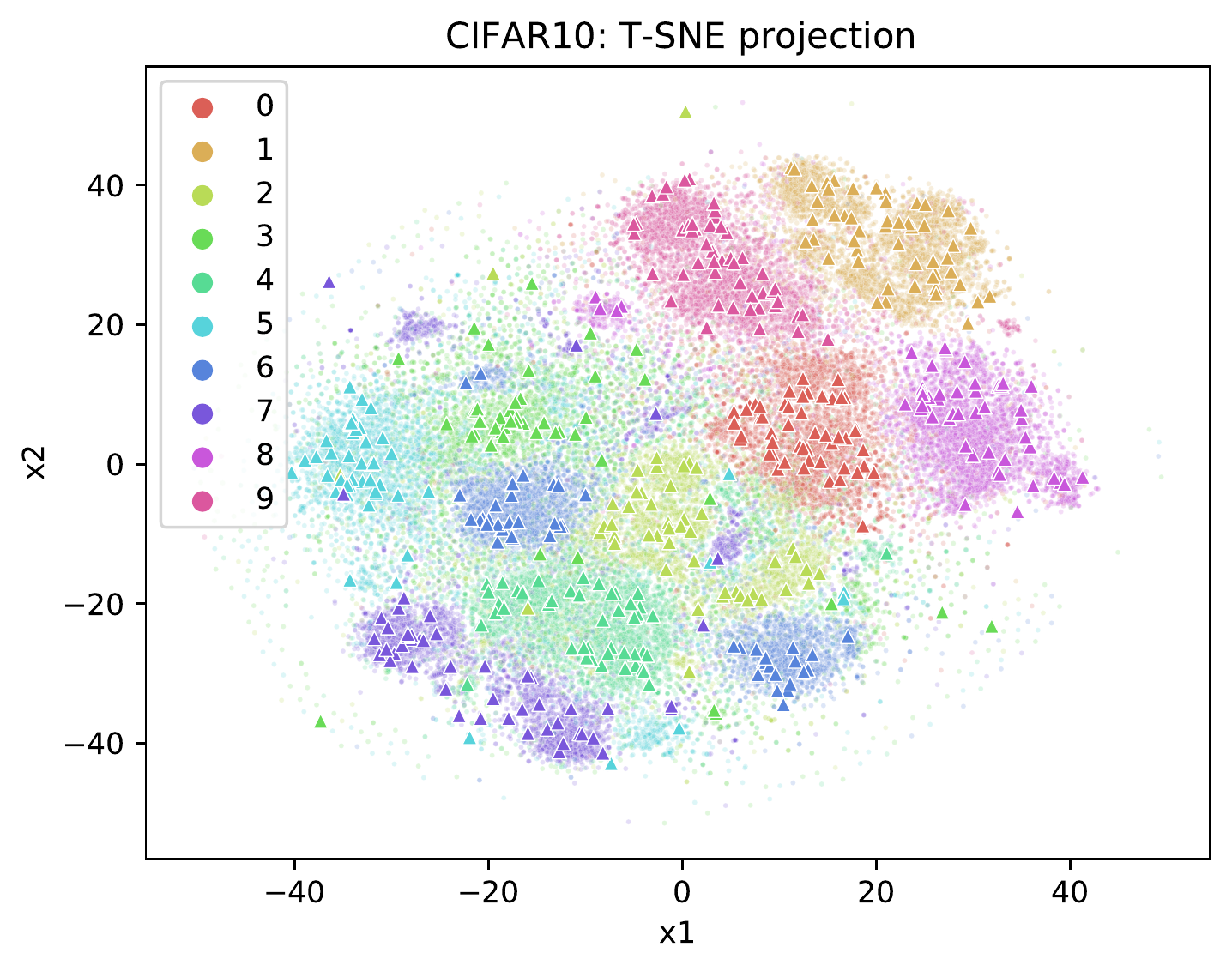}}%
\subfloat[IDC]{\label{fig:app-tsne-idc-10class} \includegraphics[width=0.5\textwidth]{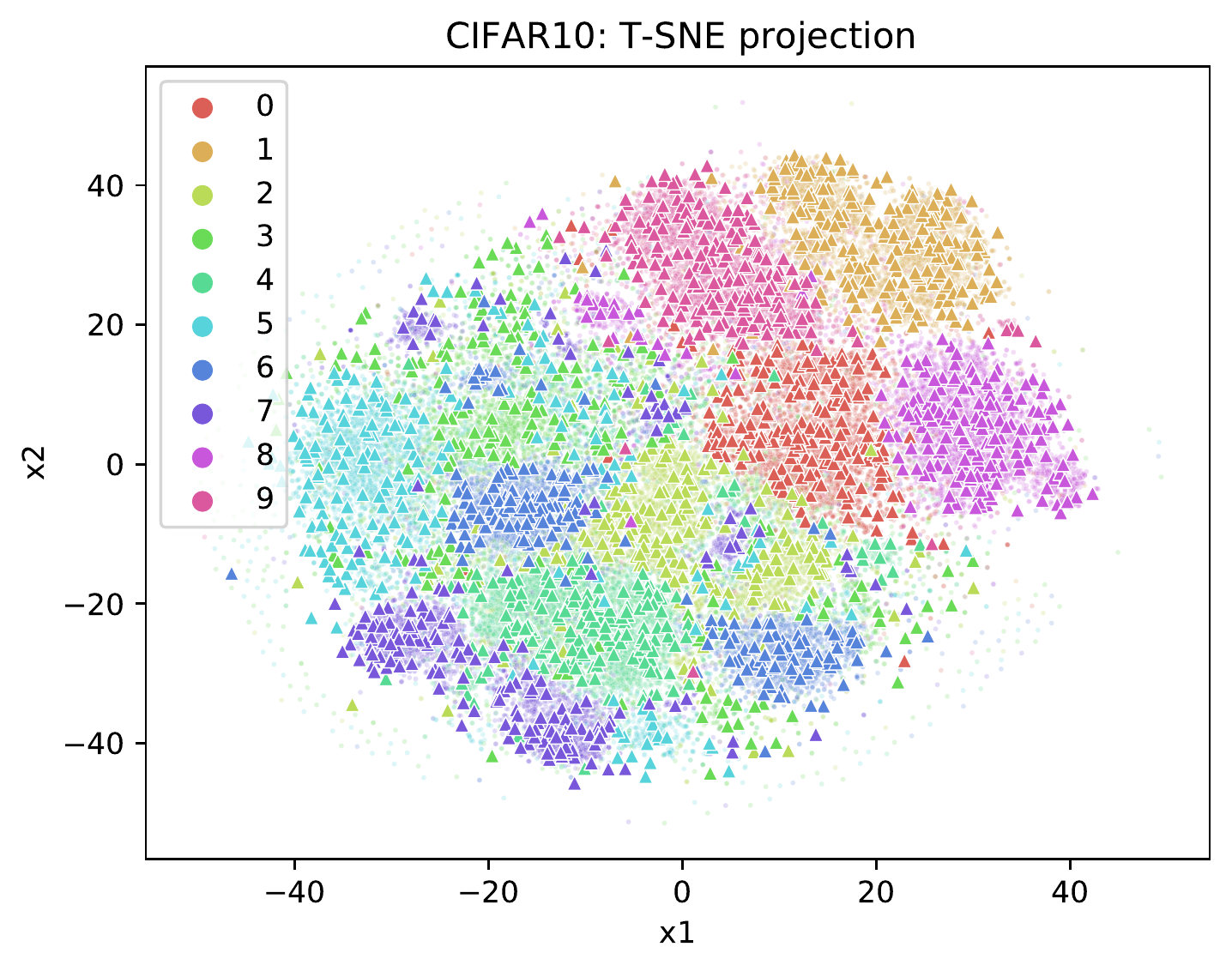}}%
\hfill
\caption{Data distribution of real images and synthetic images learned in CIFAR10 with IPC-50.}
\label{fig:app-tsne-10class}
\end{figure*}

In Figure~\ref{fig:app-tsne-10class}, we present 10-class distribution of CIFAR10 with IPC 50.
We use a pretrained ConvNet3 model to generate a 2048-dimension embedding and project that embedding into a 2-dimension plane by t-SNE with hyperparameter \textit{perplexity} 50. We see that IDC has more than 50 images per class since IDC algorithm uses a multi-formation function $f$ to decode condensed dataset into training dataset.
Visually, we find that the projected distilled images from DM, MTT, IDC can spread over the cluster better than DSA. As an example in DSA, class 2 has virtually no points at the center of the cluster. However, DM, MTT, IDC are well suited for class 2.

\bibliographystyle{named}
\bibliography{./ijcai23}

\end{document}